\documentclass[journal]{IEEEtran}
\usepackage{cite}
\usepackage{mathrsfs}
\usepackage{amsmath}
\newcommand{\norm}[1]{\left\lVert#1\right\rVert}
\usepackage{multirow}
\usepackage{algorithm}
\usepackage{algpseudocode}
\usepackage{subcaption}
\usepackage{graphicx}
\usepackage{caption}
\usepackage{subcaption}
\usepackage{amsfonts,amssymb}
\usepackage{color}
\usepackage{bm}
\usepackage{array}
\usepackage{mathtools}
\makeatletter  
\newif\if@restonecol  
\makeatother

\usepackage[algo2e,ruled,vlined]{algorithm2e}%[ruled,vlined]{  
\SetKwProg{Init}{init}{}{}
\usepackage{amsthm}
\newcommand{\argmin}{\arg \min}
\newcommand{\argmax}{\arg \max}
\usepackage{flushend}
\ifCLASSINFOpdf
\else
\fi
\usepackage{amsmath}
\begin{document}
\theoremstyle{defn}
\newtheorem{defn}{Definition}

\title{Modal Regression based Structured Low-rank\\
Matrix Recovery for Multi-view Learning}

\author{Jiamiao~Xu$^\dagger$,
        Fangzhao~Wang$^\dagger$,
        Qinmu~Peng$^*$,
        Xinge~You,~\IEEEmembership{Senior Member,~IEEE,}
        Shuo~Wang,
        \\Xiao-Yuan~Jing,
        and~C. L. Philip Chen,~\IEEEmembership{Fellow,~IEEE}
        % <-this % stops a space
\thanks{$^\dagger$The first two authors contributed equally to this work and should be
regarded as co-first authors.}
\thanks{$^*$Corresponding author.}
\thanks{J. Xu, F. Wang and S. Wang are with the School of Electronic Information
	and Communications, Huazhong University of Science and Technology,
	Wuhan 430074, China.}
\thanks{Q. Peng and X. You are both with the School of Electronic Information and Communications, Huazhong University of Science and Technology, Wuhan 430074, China and with Shenzhen Huazhong University of Science and Technology Research Institute, China.(e-mail: pqinmu@gmail.com)}
\thanks{X.-Y. Jing is with the State Key Laboratory of Software Engineering, School of Computer, Wuhan University, China.}
\thanks{C. L. P. Chen is with the Department of Computer and Information Science,
	Faculty of Science and Technology, University of Macau, Macau 99999,
	China; with Dalian Maritime University, Dalian 116026, China; and also with
	the State Key Laboratory of Management and Control for Complex Systems,
	Institute of Automation, Chinese Academy of Sciences, Beijing 100080, China.}% <-this % stops a space
}


\markboth{IEEE Transactions on Neural Networks and Learning Systems}%
{Shell \MakeLowercase{\textit{et al.}}: Bare Demo of IEEEtran.cls for IEEE Journals}



\maketitle

\begin{abstract}
Low-rank Multi-view Subspace Learning~(LMvSL) has shown great potential in cross-view classification in recent years. Despite their empirical success, existing LMvSL based methods are incapable of well handling view discrepancy and discriminancy simultaneously, which thus leads to the performance degradation when there is a large discrepancy among multi-view data. To circumvent this drawback, motivated by the block-diagonal representation learning, we propose Structured Low-rank Matrix Recovery~(SLMR), a unique method of effectively removing view discrepancy and improving discriminancy through the recovery of structured low-rank matrix. Furthermore, recent low-rank modeling provides a satisfactory solution to address data contaminated by predefined assumptions of noise distribution, such as Gaussian or Laplacian distribution. However, these models are not practical since complicated noise in practice may violate those assumptions and the distribution is generally unknown in advance. To alleviate such limitation, modal regression is elegantly incorporated into the framework of SLMR~(term it MR-SLMR). Different from previous LMvSL based methods, our MR-SLMR can handle any zero-mode noise variable that contains a wide range of noise, such as Gaussian noise, random noise and outliers. The alternating direction method of multipliers (ADMM) framework and half-quadratic theory are used to efficiently optimize MR-SLMR. Experimental results on four public databases demonstrate the superiority of MR-SLMR and its robustness to complicated noise.

\end{abstract}


% \fi
\IEEEpeerreviewmaketitle

\section{Introduction}
The rapid development of information acquisition technology endows us with the ability to describe an object in different ways, leading to multiple representations for one object. This is also known as multi-view data in the field of multi-view learning~\cite{xu2013survey,li2019a}. Although multi-view data plays an important role in improving the performance of machine learning systems in recent years~\cite{8307774,wu2016multi,zhao2018multi}, it also introduces a challenging yet significant classification problem due to the large discrepancy among views~\cite{kan2016multi,sharma2012generalized,cao2018generalized}. Particularly, it is completely impractical to directly perform classification when samples from gallery set and the ones from probe set are heterogeneous~\cite{xu2018multi}. This is often referred to as cross-view classification~\cite{sharma2012generalized}.

%Multi-view data analysis is the key to achieve more efficient machine intelligence, since data is frequently captured wih different types of sensors or taken from different viewpoints in realistic complex scene. Due to the huge gap between multi-view data~(i.e., within-class data from different views might have less similarity than between-class data from the same view), one of the most challenging issues is how to match the same object from different viewpoints or sensor, ususlly denoted as cross-view recognition. In other words, it is crucial to mitigate the view divergence.

Substantial efforts have been made to tackle this problem in recent years. One of the most representative methods is based on the Multi-view Subspace Learning~(MvSL)~\cite{nielsen2002multiset,chaudhuri2009multi,hotelling1936relations} that aims to learn multiple view-specific mapping functions to project multi-view data into a common subspace, in which the view discrepancy can be removed. Canonical Correlation Analysis~(CCA)~\cite{hotelling1936relations,thompson2005canonical} and Partial Least Square~(PLS)~\cite{5995350,cai2013regularized} are two earliest MvSL based approaches that devote to decrease view discrepancy by maximizing correlation or convariance of projected samples. The best-known supervised methods, Generalized Multi-view Analysis~(GMA)~\cite{sharma2012generalized} and Multi-view Discriminant Analysis~(MvDA)~\cite{kan2016multi}, were later proposed to further incorporate intra-view or inter-view supervised information, thus improving classification performance. Multi-view Hybrid Embedding~(MvHE)~\cite{xu2018multi} was recently proposed to take into consideration view discrepancy, discriminancy and nonlinearity simultaneously with a divide-and-conquer strategy. Moreover, some deep neural networks~(DNN) based models~\cite{andrew2013deep,wang2015deep,kan2016multideep} are proposed to extract nonlinear features with stronger representation power. It is worth mentioning that some work in other related areas also provided feasible solutions, such as low-rank coding~\cite{ding2018deep}, Gaussian Process Latent Variable Model (GPLVM)~\cite{song2015similarity} or Maximum Mean Discrepancy (MMD)~\cite{li2018transfer,li2018heterogeneous}. Despite promising results on real applications, aforementioned MvSL based methods will fail to work, when the view information of test samples is unknown in advance~\cite{ding2014low}. %Another is worth mentioning that aforementioned work except for MvDA is incapable of handling incomplete-view data~\cite{xu2015multi}, whereas one example is likely to be missing its representation on one view in practice. 

Low-rank Multi-view Subspace Learning (LMvSL) was later proposed to learn a common projection for all views via low-rank reconstruction so that a newly coming test sample can be directly mapped into a subspace even if we do not know which view it comes from. One of the most well-known unsupervised LMvSL based methods is Low-rank Common Subspace (LRCS) \cite{ding2014low} that attempts to find a common subspace, in which projected samples have a low-rank representation with respect to a given dictionary. Although the low-rank property helps to partly decrease view discrepancy and improve discriminancy, the neglect of discriminant information prevents LRCS from achieving a better classification performance. Low-rank Discriminant Embedding~(LRDE)~\cite{Li2017Low} was thereafter proposed to further sufficiently use discriminant information under the framework of graph embedding. One should note that both LRCS and LRDE prefer to learn a low-rank common projection to recover shared information among views. However, the nuclear norm on projection matrix also leads to much redundant information, which is not good for classification. To learn discriminant features with less redundancy, Collective Low-rank Subspace~(CLRS)~\cite{ding2018robust} attempts to learn a full-rank projection and extends LRCS to a supervised method via Fisher criterion. Furthermore, Robust Multi-view Subspace Learning~(RMSL)~\cite{Ding:2016:RMS:3015812.3015987} was proposed to learn common subspace through dual low-rank decomposition to uncover class structure and view-variance structure behind multi-view data. Benefiting from the utilization of graph embedding, LRDE, CLRS and RMSL achieve more satisfactory performance than LRCS. Albeit the simplicity of graph embedding, the similarity of samples from different views cannot be precisely measured due to the large discrepancy among views, thus resulting in an inaccurate weight matrix of graph~\cite{xu2018multi,YOU201937}. Considering that multi-view data is more likely to be corrupted~\cite{li2018multi}, LMvSL based methods introduce an error term with certain regularization for modeling noise, and these commonly used regularizations work for one predefined assumption on the distribution of the noise variable. However, complicated noise in practice may violate those assumptions and its distribution is generally unknown in advance. Consequently, there still remains a need for a LMvSL based method that can better utilize supervised information and handle various noise.

This work aims at solving above two problems. To consider discriminant information while circumventing the limitation of graph embedding, different from previous work, this paper proposes a novel SLMR algorithm, which incorporates structured regularization into the fundamental framework of LMvSL based approaches to learn a multiple block-diagonal representation. Consequently, the view discrepancy is removed and inter-view discriminancy is improved. Moreover, inspired by the success of modal regression in handling complicated noise~\cite{wang2017cognitive,feng2017statistical}, MR-SLMR is thereafter proposed by integrating modal regression into SLMR, which thus leads to a more robust model.

%Previous work on dictionary learning~\cite{zhang2013learning,zhang2018discriminative,xie2018implicit,lu2019subspace} indicates that block-diagonal structure for representation can effectively capture underlying discriminative information. Motivated by this idea, we propose a novel LMvSL based algorithm, which imposes normalization, non-negative and discriminative bound term on representation of each view to learn multiple block-diagonal representations and attempts to learn a lowest-rank structure  by imposing low-rank constraint on the joint matrix of multiple block-diagonal representations. Specifically, the block-diagonal structure of each view aims to improve intra-view discriminancy, whereas the learning of lowest-rank structure for joint representation helps to decrease view discrepancy,improve inter-view discriminancy and further enhance intra-view discriminancy. Furthermore, to improve the robustness of our method to more complicated noise, inspired by the success of modal regression, we flexibly integrate modal regression into our algorithm. 

To summarize, our main contributions are threefold:

1) A novel algorithm, named SLMR, is proposed for cross-view classification to properly take advantage of discriminant information.

2) As a regression technique, modal regression is elegantly incorporated into SLMR to enable our method to flexibly deal with various noise. Furthermore, an alternating algorithm is designed to efficiently optimize MR-SLMR.

3) Extensive experiments conducted on four public datasets demonstrate that our MR-SLMR is able to take into account supervised information more effectively than other methods based on graph embedding, thus boosting the classification performance. Moreover, compared with LMvSL based counterparts, our method is more robust to complicated noise.

The remainder of this paper is organized as follows. Sect.~\ref{sec:related_work} introduces the related work on LMvSL and modal regression. Sect.~\ref{sec:proposed_method} presents the formulation of our method, its optimization, complexity analysis and relevant discussion on MR-SLMR. Experiments are conducted in Sect.~\ref{sec:experiments}. Finally, Sect.~\ref{sec:conclusion} concludes this paper.

\section{Related Work}
\label{sec:related_work}
In this section, key notations used throughout this paper are summarized and the relevant Low-rank Multi-view Subspace Learning~(LMvSL) based methods are briefly reviewed. Furthermore, some knowledge on modal regression is introduced to make readers easier to understand our work.

\subsection{Notation}  % the definition of l_0 and l_2,0
\label{sec:notation}
Scalars are represented by lowercase letters~(e.g., $x$), vectors and matrices are denoted by bold lowercase letter~(e.g., $\bm{x}$) and bold uppercase letter~(e.g., $\bm{X}$) respectively. Moreover, it is worth mentioning that random variable used in Sect.~\ref{subsec:modal_regression} is also denoted by bold uppercase letter and its value appears as bold lowercase letter. Let $\bm{\mathrm{I}}$ denote identity matrix, $\bm{X}^\mathrm{T}$ denote the transpose of matrix $\bm{X}$ and $\mathrm{tr}\left(\cdot\right)$ denote the trace operator. Matrix norms, $\norm{\bm{X}}_{F}$, $\norm{\bm{X}}_{0}$, $\norm{\bm{X}}_{2,0}$ and  $\norm{\bm{X}}_{\infty}$, are frequently used in this paper, and these norms are defined as $\norm{\bm{X}}_F\!=\!\sqrt{\sum_i\sum_j|\bm{X}_{ij}|^2}$,
$\norm{\bm{X}}_{0}\!=\!\#\{\left(i,j\right):\bm{X}_{ij}\!\neq\!0\}$, 
$\norm{\bm{X}}_{2,0}\!=\!\#\{j:\sqrt{\sum_{i}(\bm{X}_{ij})^2}\!\neq\!0\}$ and $\norm{\bm{X}}_{\infty}\!=\!\max{\{|\bm{X}_{ij}|}\}$ respectively, where $\bm{X}_{ij}$ denotes the $i$-th row $j$-th column element of $\bm{X}$, $\left\{\cdot\right\}$ denotes a set symbol and $\#\{\cdot\}$ stands for the number of elements in the set. In practice, $\norm{\bm{X}}_{0}$ and $\norm{\bm{X}}_{2,0}$ are generally replaced with $\norm{\bm{X}}_{1}\!=\!\sum_{ij}|\bm{X}_{ij}|$ and $\norm{\bm{X}}_{2,1}\!=\!\sum_{j}\sqrt{\sum_{i}(\bm{X}_{ij})^2}$ to facilitate optimization. Furthermore, $\norm{\bm{X}}_*$ stands for nuclear norm, i.e., the sum of singular values of matrix.    

\subsection{Low-rank Multi-view Subspace Learning based Approaches}
\label{sec:lmvsl_methods}
Suppose $\bm{X}\!=\![\bm{X}_1,\bm{X}_2,\cdots,\bm{X}_k]\!\in\!\mathbb{R}^{d\!\times\!m}$~($k\geq2$) denotes multi-view data collected from $k$ domains, where $m$ is the number of samples, $d$ stands for feature dimensionality and $\bm{X}_v$ denotes the data matrix from the $v$-th view. LMvSL based methods attempt to learn a common mapping function $\bm{P}$ shared by all views to project $\bm{X}$ into a $p$-dimensional subspace, where the discrepancy among views can be reduced. Its general objective can be formulated as below:
\begin{align}\label{basic_lrr_msl}
\min_{\bm{Z},\bm{E},\bm{P}}&\mathrm{rank}\left(\bm{Z}\right)\!+\!\lambda\norm{\bm{E}}_\ell\nonumber\\
s.t.\quad &\bm{P}^\mathrm{T}\bm{X}\!=\!\bm{A}\bm{Z}+\bm{E},
\end{align}
where $\bm{Z}$ is the low-rank representation of $\bm{P}^\mathrm{T}\bm{X}$ with respect to an over-complete dictionary $\bm{A}$ and $\lambda$ is a trade-off parameter. LMvSL based methods introduce error term $\bm{E}$ to model noise, where the regularization strategy is determined by the noise type, such as $\norm{\bm{E}}_1$ for random noise~\cite{wright2009robust,donoho2006most,candes2006stable}, $\norm{\bm{E}}_F$ for Gaussian noise~\cite{pearson1901liii,candes2010matrix} and $\norm{\bm{E}}_{2,1}$ for outliers~\cite{liu2013robust}.

Problem~(\ref{basic_lrr_msl}) is hard to optimize due to the discrete nature of the rank function~\cite{fazel2004rank}. A common practice to handle this problem is to replace rank function with nuclear norm~\cite{liu2013robust,li2016learning}, and then Eq.~(\ref{basic_lrr_msl}) can be substituted with the following convex optimization~\cite{candes2011robust,candes2009exact}:
\begin{align}\label{basic_lrr_msl2}
\min_{\bm{Z},\bm{E},\bm{P}}&\norm{\bm{Z}}_*\!+\!\lambda\norm{\bm{E}}_\ell\nonumber\\
s.t.\quad &\bm{P}^\mathrm{T}\bm{X}\!=\!\bm{A}\bm{Z}+\bm{E}.
\end{align}
Obviously, within-class samples present the same representation in the common subspace with a higher probability when a low-rank $\bm{Z}$ is learned by optimizing Eq.~(\ref{basic_lrr_msl2}), which thus removes the view discrepancy and improves discriminability.
%LMvSL based methods assume that within-class samples from all views have the same representation vector with a high probability when $\bm{Z}$ is low rank, thus removing the discrepancy among views and improving discriminability.

%More efforts on LMvSL has been made in recent years due to the invalidation of assumption on some real applications. We introduce the most related LMvSL base methods below:
More efforts on LMvSL have been taken in recent years. We then analyze these work below.

\subsubsection{LRCS}
Low-rank Common Subspace Learning (LRCS) \cite{ding2014low} aims to learn a common projection $\bm{P}$ and multiple view-specific ones $\{\bm{E}_v\}_{v=1}^k$ to discover hidden structure of multi-view data. Its objective can be formulated as:
\begin{align}\label{LRCS}
&\min_{\bm{Z},\bm{E},\bm{P},\bm{P}_i,\bm{E}_p}\norm{\bm{Z}}_*\!+\!\norm{\bm{P}}_*\!+\!\lambda_{1}\norm{\bm{E}}_{2,1}\!+\!\lambda_{2}\norm{\bm{E}_P}_1\nonumber\\
&s.t.\quad \left[\bm{P}_{1}^\mathrm{T}\bm{X}_1,\dots,\bm{P}_{k}^\mathrm{T}\bm{X}_k\right]\!=\!\bm{A}\bm{Z}+\bm{E},\nonumber\\
&\bm{P}_S\!=\!\left[\bm{P}_1,\bm{P}_2,\ldots,\bm{P}_k\right],\quad \bm{P}_T\!=\!\left[\bm{P},\bm{P},\ldots,\bm{P}\right],\nonumber\\
&\bm{P}^\mathrm{T}\bm{P}\!=\!\bm{\mathrm{I}},\quad \bm{P}_S\!=\!\bm{P}_T\!+\!\bm{E}_{P},\quad \bm{E}_P\!=\!\left[\bm{E}_1,\bm{E}_2,\ldots\bm{E}_k\right].
\end{align}

Note that LRCS suffers from the neglect of discriminant information.

\subsubsection{LRDE}
Low-rank Discriminant Embedding~(LRDE)~\cite{Li2017Low} was later proposed to learn discriminant common subspace by utilizing supervised information under the well-known framework of graph embedding:
\begin{align}\label{}
\min_{\bm{Z},\bm{E},\bm{P}}&\norm{\bm{Z}}_*\!+\!\lambda_{1}\norm{\bm{P}}_*\!+\!\lambda_{2}\norm{\bm{E}}_{2,1}\!+\!\lambda_{3}\mathrm{tr}\left(\bm{P}^\mathrm{T}\bm{X}\bm{L}_w\bm{X}^\mathrm{T}\bm{P}\right)\!\nonumber\\
s.t.\quad &\bm{P}^\mathrm{T}\bm{X}\!=\!\bm{A}\bm{Z}+\bm{E},\quad \bm{P}^\mathrm{T}\bm{X}\bm{L}_b\bm{X}^\mathrm{T}\bm{P}\!=\!\bm{\mathrm{I}},
\end{align}
% where $\bm{L}_w$ and $\bm{L}_b$ are the graph Laplacian of within-class samples and between-class samples~\cite{?,?}, respectively. %These sentence needs to be check again!
where $\bm{L}_w$ and $\bm{L}_b$ are the graph Laplacian matrix. Note that criterion A and criterion B in~\cite{Li2017Low} with respect to $\bm{L}_w$ and $\bm{L}_b$ are used to guide the construction of weight matrices, where the former intends to connect within-class nearest sample pairs, whereas the latter aims to connect intra-view between-class nearest sample pairs.
 
 %\textcolor{red}{The former connects nearest sample pairs of same class, which describe the transition of view in the same class~(criterion A in \cite{Li2017Low}). And the latter connects intra-view nearest sample pair of different class, which characterize the itra-view discrimination~(criterion B in \cite{Li2017Low})}.
\subsubsection{CLRS}
A similar method to LRDE was the recently proposed Collective Low-rank Subspace~(CLRS)~\cite{ding2018robust} that extends LRCS to a supervised model with a graph regularizer:
\begin{align}\label{CLRS}
&\min_{\bm{Z},\bm{E},\bm{P},\bm{P}_i,\bm{S}_i}\norm{\bm{Z}}_*\!+\!\lambda_{1}\norm{\bm{E}}_{2,1}+\!\lambda_{2}\sum_{i=1}^{k}\norm{\bm{S}_i}_1\!+\!\Omega\left(\bm{P},\bm{Z}\right)\nonumber\\
&s.t.\quad [\bm{P}_{1}^\mathrm{T}\bm{X}_1,\dots,\bm{P}_{k}^\mathrm{T}\bm{X}_k]\!=\!\bm{A}\bm{Z}+\bm{E},\nonumber\\
&\quad\quad \bm{P}^\mathrm{T}\bm{P}\!=\!\bm{\mathrm{I}},\quad \bm{P}_i\!=\!\bm{P}\!+\!\bm{S}_i,\quad i\!=\!1,\dots,k,
\end{align}
where regularization term $\Omega\left(\bm{P},\bm{Z}\right)$ is used to incorporate discriminant information, and orthogonal constraint on $\bm{P}$ is to reduce redundant information~\cite{li2016learning}.

\subsubsection{RMSL}
Robust Mutli-view Subspace Learning~(RMSL) \cite{Ding:2016:RMS:3015812.3015987} attempts to decompose global low-rank structure into view structure and class structure:
\begin{align}\label{}
\min_{\bm{Z}_v,\bm{Z}_c,\bm{E},\bm{P}}&\norm{\bm{Z}_v}_*\!+\!\norm{\bm{Z}_c}_*\!+\!\lambda_{1}\norm{\bm{E}}_{1}\!+\!\mathcal{G}_c(\bm{P},\bm{Z}_c)\!+\!\mathcal{G}_v(\bm{P},\bm{Z}_v)\!\nonumber\\
s.t.\quad &\bm{P}^\mathrm{T}\bm{X}\!=\!\bm{A}(\bm{Z}_v+\bm{Z}_c)+\bm{E},\quad \bm{P}^\mathrm{T}\bm{P}\!=\!\bm{\mathrm{I}},
\end{align}
where graph regularizers $\mathcal{G}_c$ and $\mathcal{G}_v$ integrate supervised information into class structure and view structure respectively.

\begin{figure*}
    \centering
    \begin{minipage}[t]{7.3cm}
    \centering
		\includegraphics[width=7.3cm]{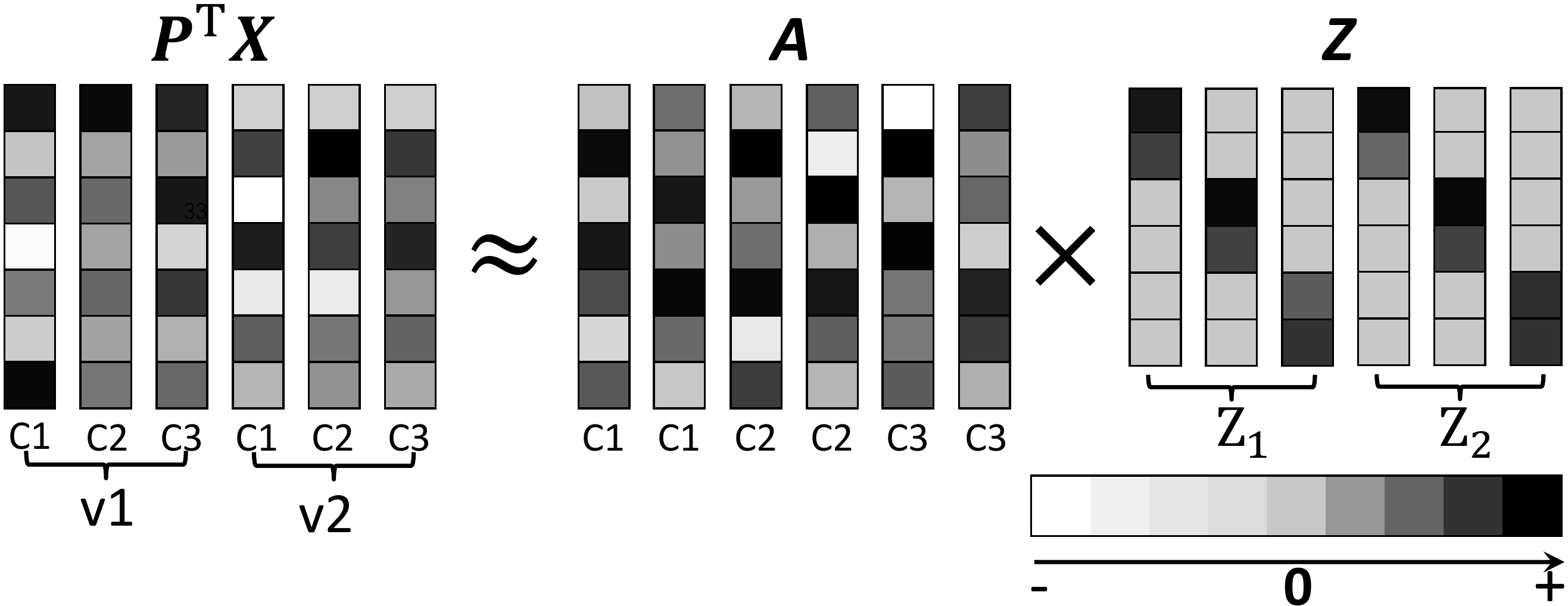}
		\subcaption{}
	\end{minipage}
	\hspace{0.6cm}
    \begin{minipage}[t]{3.2cm}
    \centering
		\includegraphics[width=3.2cm]{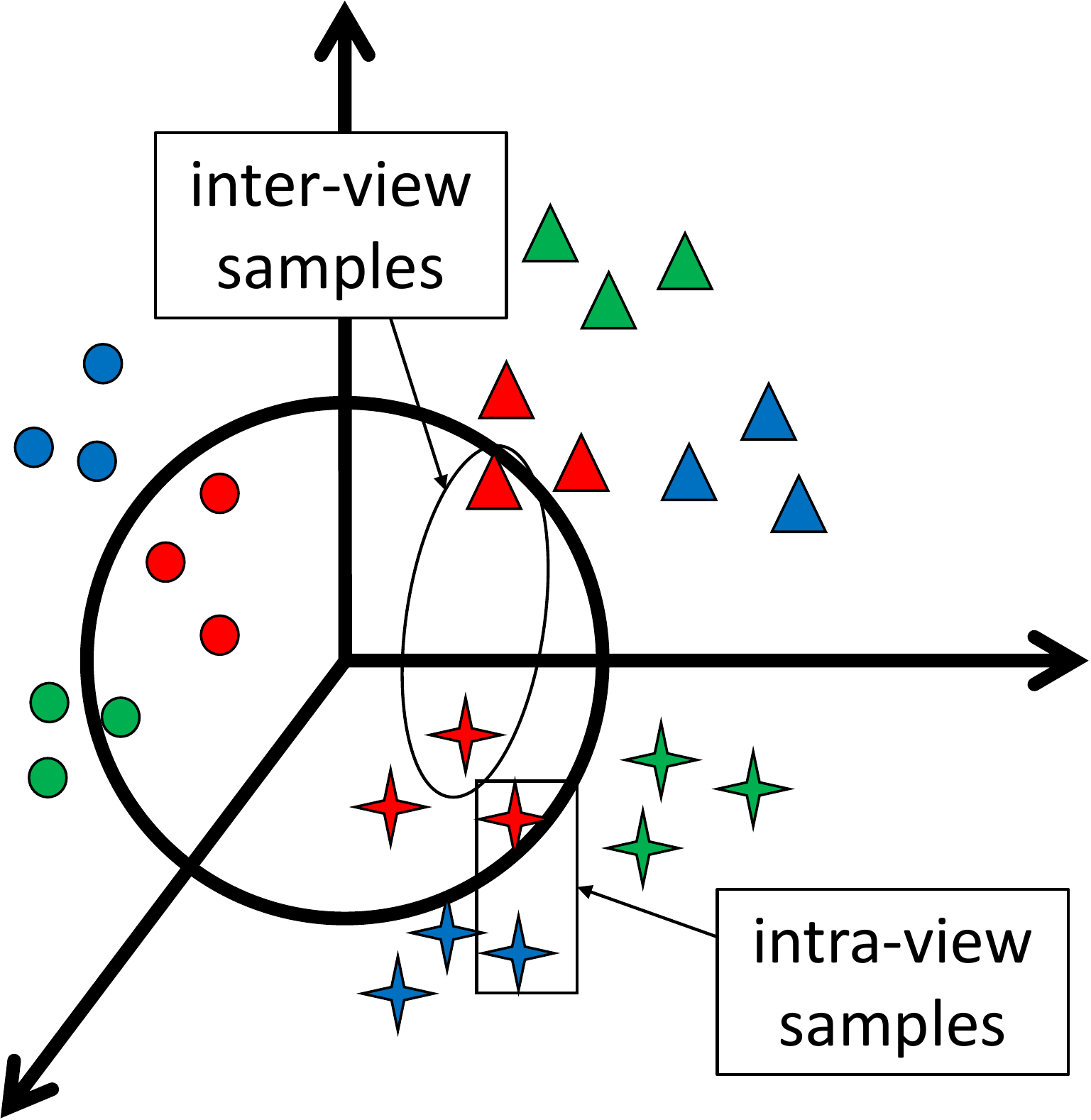}
		\subcaption{}
	\end{minipage}
	\begin{minipage}[t]{3.2cm}
	\centering
		\includegraphics[width=3.2cm]{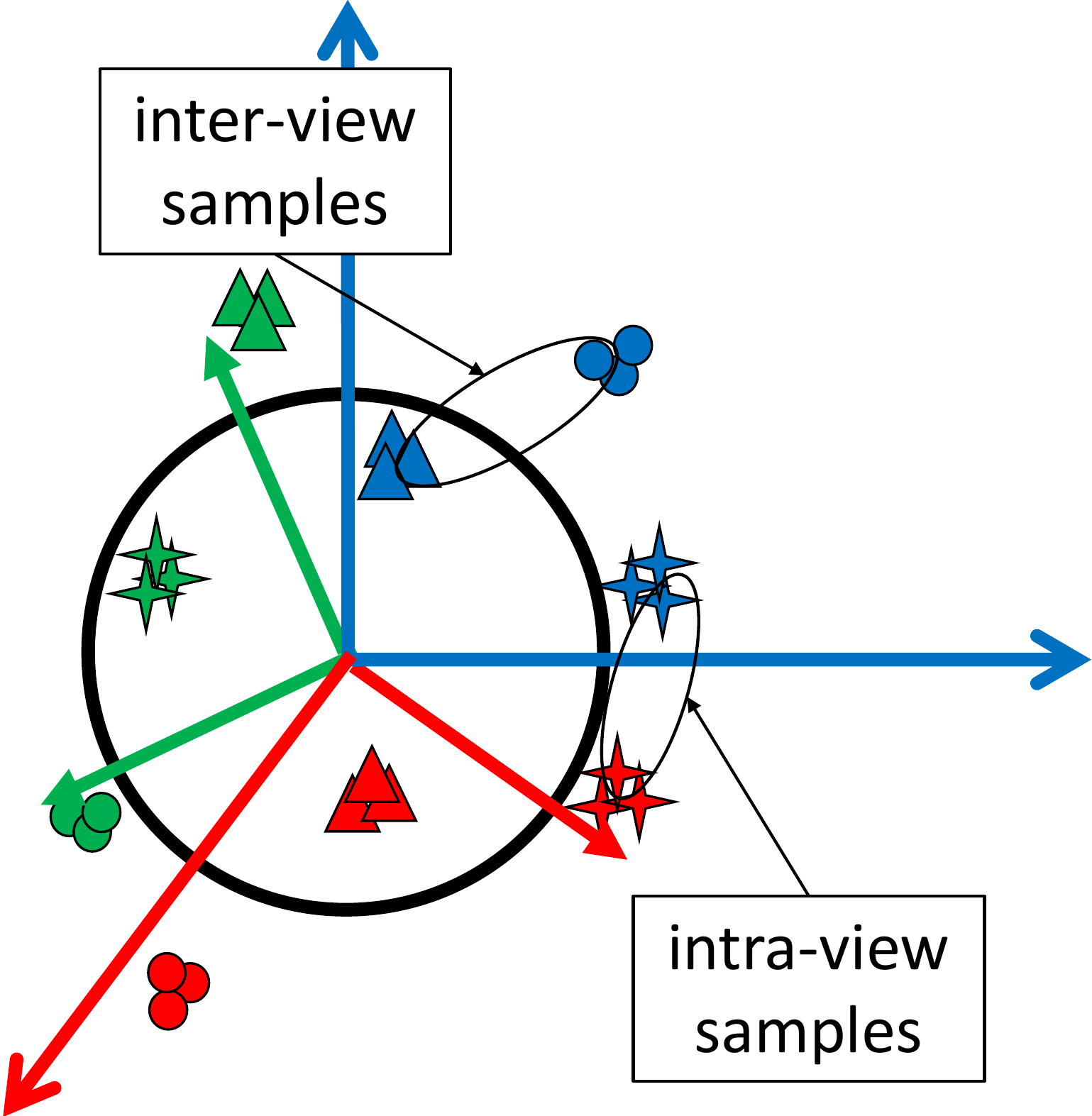}
		\subcaption{}
	\end{minipage}
	\begin{minipage}[t]{3.2cm}
	\centering
		\includegraphics[width=3.2cm]{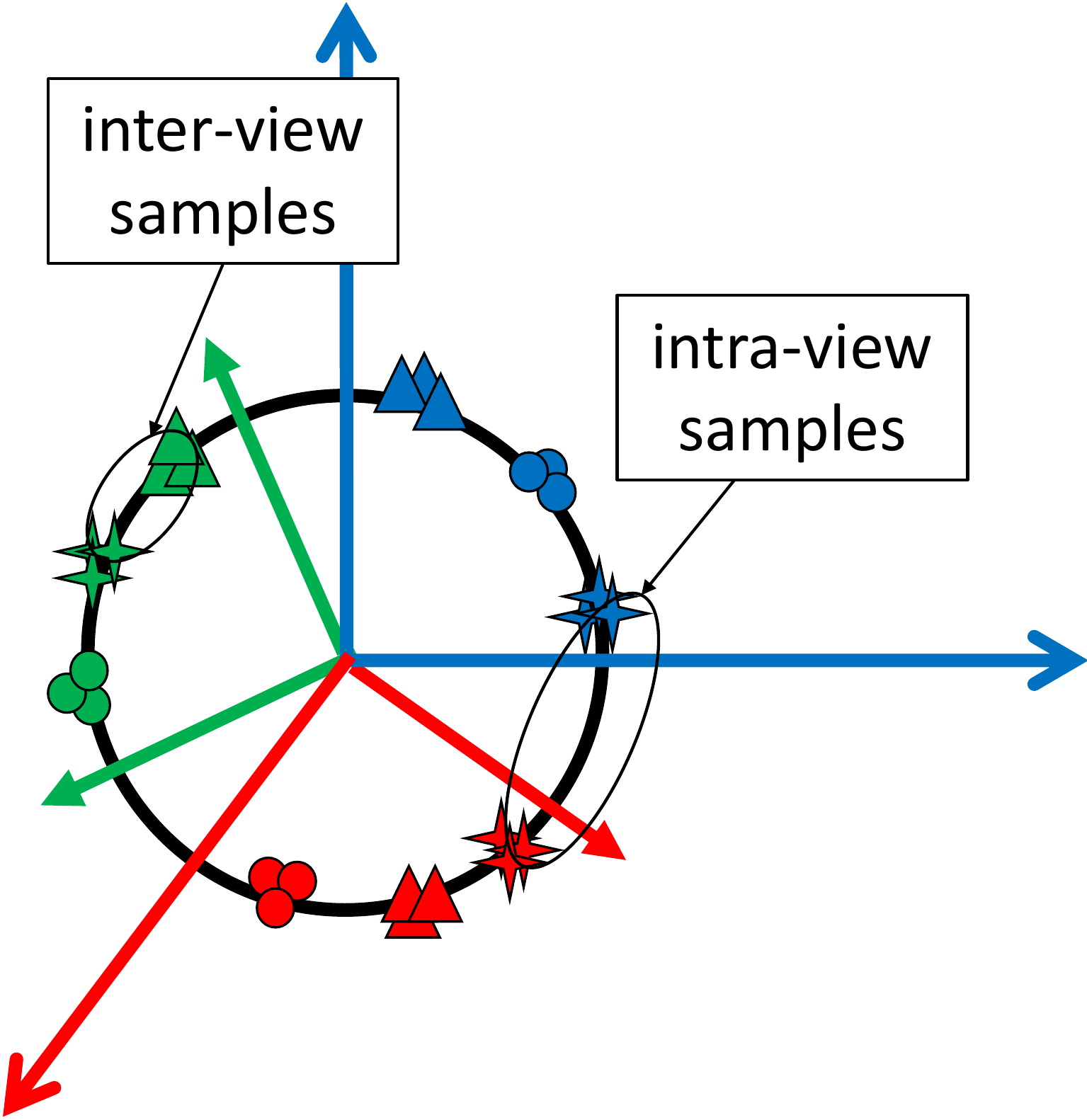}
		\subcaption{}
	\end{minipage}
    \caption{An overview of SLMR. As shown in (a), SLMR aims to learn multiple block-diagonal matrix, one for each view~(i.e., $\bm{Z}_1$ for view $1$ and $\bm{Z}_2$ for view $2$). Shapes shown in (b), (c) and (d) are representation vectors generated by various learning strategies. Markers and colors shown in (b), (c) and (d) stand for views and classes respectively. Coordinate axes with colors in (c) and (d) denote the bases of different classes from the dictionary. We employ various strategies to learn representation matrix $\bm{Z}$ during the process of dictionary learning. Then, we can learn a subspace of $\bm{Z}$ shown in (b) when commonly used regularization (e.g., sparse representation) is imposed on $\bm{Z}$ and we can learn a subspace of $\bm{Z}$ shown in (c) when nuclear norm is imposed on $\bm{Z}$. Moreover, we can learn a subspace of $\bm{Z}$ shown in (d) when nuclear norm and structured regularization are both imposed on $\bm{Z}$.
    \vspace{0.0cm}}
    \label{fig:SLMR}
\end{figure*}

\subsection{Modal Regression}
\label{subsec:modal_regression}
Despite great progress in handling noise, current regularizations require noise to follow a specific predefined distribution (e.g., $\ell_1$ for Laplacian distribution and $\ell_2$ for Gaussian distribution). To alleviate such limitation, a more general regularization is expected to be designed. A promising solution is the modal regression~\cite{sager1982maximum} that has been proved to be insensitive to zero-mode noise theoretically and practically~\cite{feng2017statistical,wang2017cognitive}. Before elaborating our method, we introduce basic knowledge on modal regression below.

Modal regression intends to estimate the conditional mode of response random variable $\bm{Y}\!\in\!\mathcal{Y}$ given input $\bm{X}\!=\!\bm{x}\!\in\!\mathcal{X}$, where $\bm{Y}\in\mathbb{R}$, $\bm{X}\in\mathbb{R}^d$ and mode refers to the value that appears most frequently. Suppose that i.i.d observations $\bm{z}\!=\!\{\left(\bm{x}_i,\bm{y}_i\right)\}_{i=1}^m\!\subset\!\mathcal{X}\times\mathcal{Y}$ are generated by:
\begin{align}\label{modal_regression}
    \bm{Y}\!=\!f^*\left(\bm{X}\right) + \bm{N},
\end{align}
where $f^*$ is the unknown target function and $\bm{N}$ is the noise term. Modal regression aims to obtain the target function $f^*\left(\bm{X}\right)$ via the following modal regression function $f_M$~\cite{collomb1986note}:
\begin{align}
    f_M\left(\bm{x}\right)\!\coloneqq\!\argmax_{t}p_{\bm{Y}|\bm{X}}\left(t|\bm{X} = \bm{x}\right),
\end{align}
where $p_{\bm{Y}|\bm{X}}\left(t|\bm{X}=\bm{x}\right)$ is the conditional probability density of $\bm{Y}$ conditioned on $\bm{X}$. Note that we have $f^*\left(\bm{X}\right)\!=\!f_M\left(\bm{X}\right)$ according to Eq.~(\ref{modal_regression}) when the mode of noise $\bm{N}$ at any $\bm{x}$ is zero~\cite{feng2017statistical}. Hence, under the zero-mode noise assumption, the target function $f^*\left(\bm{X}\right)$ can be obtained by estimating the modal regression function $f_M\left(\bm{X}\right)$.

To better estimate $f_M\left(\bm{X}\right)$, modal regression risk $\mathcal{R}\left(f\right)$ \cite{feng2017statistical} is introduced:

\begin{align}
    \mathcal{R}\left(f\right)\!=\!\int_{\mathbb{R}^d}p_{\bm{Y}|\bm{X}}\left(f(\bm{x}\right)|\bm{X}=\bm{x})d\rho_{\bm{X}}\left(\bm{x}\right),
\end{align}
where $\rho_{\bm{X}}$ is the marginal distribution of $\bm{X}$. Then, we have $f_M\left(\bm{X}\right)\!=\!\argmax_{f\in\mathcal{F}}{R\left(f\right)}$ according to~\cite{feng2017statistical}. 

Let $\bm{E}_f\!=\!\bm{Y}\!-\!f\left(\bm{X}\right)$ denote the error random variable for the measure function $f\left(\bm{X}\right)$. Then the density of $\bm{E}_f$ can be written as:
\begin{align}
    p_{\bm{E}_f}\left(e_f\right) = \int_{\mathbb{R}^d}p_{\bm{E}_f|\bm{X}}\left(e_f|\bm{X}=\bm{x}\right)d\rho_{\bm{X}}\left(\bm{x}\right).
\end{align}
Theorem 5.1 in \cite{feng2017statistical} indicates that $\mathcal{R}\left(f\right)\!=\!p_{\bm{E}_f}\left(0\right)$. Under the zero-mode assumption, we have:
\begin{align}\label{prob_e}
    f^* = f_M = \argmax_{f\in\mathcal{F}}{p_{\bm{E}_f}\left(0\right)}.
\end{align}
Obviously, the solution of target function $f^*$ is converted to the estimation of $p_{\bm{E}_f}\left(e_f\right)$. Due to the limited space, the detailed estimation methods can refer to~\cite{wang2017cognitive,he2011maximum} or Sect.~\ref{subsec:optimization}.

\section{Modal Regression based Structured Low-rank Matrix Recovery}
\label{sec:proposed_method}
In this section, we first detail the general idea of our MR-SLMR, and then present relevant discussion and optimization. The complexity analysis is finally conducted. 

\subsection{Structured Low-rank Matrix Recovery}
Suppose we are given data $\bm{X}=\left[\bm{X}_1,\bm{X}_2,\dots,\bm{X}_k\right]\!\in\!\mathbb{R}^{d\times m}$ from $k$ views, where $d$ is the feature dimensionality, $m$ is the number of samples, $\bm{X}_v\!\in\!\mathbb{R}^{d\times m_v}$ is the $v$-th view data, containing $C$ classes, and $m_v$ is the number of samples from view $v$. 

MvSL based methods~\cite{sharma2012generalized,kan2016multi,xu2018multi} are designed under a full-view assumption~\cite{xu2015multi}. Nevertheless, more than one example could be missing its representation on one view for reasons, such as sensors failure or man-made faults~\cite{xu2015multi}. Fortunately, dictionary learning provides a feasible solution to tackle incomplete-view problem by reason of its no-dependency on paired samples~\cite{sharma2012generalized,xu2018multi}. To this end, we represent projected samples under an over-complete dictionary $\bm{A}$:
\begin{align}\label{obj:dictionary}
\min_{\bm{Z},\bm{E},\bm{P}}&\norm{\bm{Z}}_1\!+\!\lambda_{1}\norm{\bm{E}}_\ell\nonumber\\
s.t.\quad &\bm{P}^\mathrm{T}\bm{X}\!=\!\bm{A}\bm{Z}+\bm{E},
\end{align}
where $\bm{Z}$ and $\bm{E}$ are representation matrix and error matrix respectively. Then, a subspace in Fig.~\ref{fig:SLMR}(b) can be learned by optimizing Eq.~(\ref{obj:dictionary}).

To further remove view discrepancy of representation vectors~(see Fig.~\ref{fig:SLMR}(b)), different from sparse representation that focuses on each sample independently, low-rank representation provides an elegant manner to establish relations among samples~\cite{HeRobust, liu2010robust}. To this end, Eq.~(\ref{obj:dictionary}) can be rewritten as Eq.~(\ref{basic_lrr_msl2}).
% \begin{align}\label{obj:low-rank}
% \min_{\bm{Z},\bm{E},\bm{P}}&\norm{\bm{Z}}_{*}\!+\!\lambda_{1}\norm{\bm{E}}_\ell\nonumber\\
% s.t.\quad &\bm{P}^\mathrm{T}\bm{X}\!=\!\bm{A}\bm{Z}+\bm{E}.
% \end{align}

It can be expected that a subspace in Fig.~\ref{fig:SLMR}(c) can be learned by minimizing Eq.~(\ref{basic_lrr_msl2}). As shown, the view discrepancy is partially removed and inter-view discriminancy is improved.

View discrepancy can be further decreased and inter-view discriminancy can be further enhanced when the representation vector of each sample lies in the space spanned by within-class bases from dictionary~(see Fig.~\ref{fig:SLMR}(d)). Motivated by this idea, we first define $\bm{L}=\left[\bm{l}_1,\bm{l}_2,\ldots,\bm{l}_m\right]$ as the label indicator matrix of training samples, where if a sample $\bm{x}$ is from $i$-th class, its label indicator vector $\bm{l}$ is defined as:
\begin{align}
\bm{l}=\left[\overbrace{0,...,0}^{i-1},1,\overbrace{0,...,0}^{C-i}\right]^\mathrm{T}.
\end{align}
Then, a structured regularization is integrated into Eq.~(\ref{basic_lrr_msl2}) to learn a multiple block-diagonal structure of $\bm{Z}$ as follows:
\begin{align}\label{obj:struct1}
&\min_{\bm{Z},\bm{E},\bm{P}}\norm{\bm{Z}}_{*}\!+\!\lambda_{1}\norm{\bm{E}}_\ell\nonumber\\
&s.t.\quad \bm{P}^\mathrm{T}\bm{X}\!=\!\bm{A}\bm{Z}+\bm{E},~\bm{L}_s\bm{Z}\!=\!\bm{L},~\bm{Z}\!\geq\!0,
\end{align}
where $\bm{L}_s$ is the label indicator matrix of dictionary $\bm{A}$. Obviously, the representation vector for each projected sample is a linear combination of within-class bases from dictionary. 

To strengthen the generalization ability, an error term $\bm{E}_L$ is introduced. Furthermore, similar to~\cite{ding2018robust,Li2017Low}, we replace $\bm{A}$ with $\bm{P}^\mathrm{T}\bm{X}$ for simplicity, and we have $\bm{L_s}=\bm{L}$. Then, Eq.~(\ref{obj:struct1}) can be rewritten as
\begin{align}\label{obj:struct2}
&\min_{\bm{Z},\bm{E},\bm{E}_L,\bm{P}}\norm{\bm{Z}}_{*}\!+\!\lambda_{1}\norm{\bm{E}}_\ell+\lambda_{2}\norm{\bm{E}_L}_F^2\nonumber\\
&s.t.\quad \bm{P}^\mathrm{T}\bm{X}\!=\!\bm{P}^\mathrm{T}\bm{X}\bm{Z}+\bm{E},~\bm{L}\bm{Z}\!=\!\bm{L}+\bm{E}_L,~\bm{Z}\!\geq\!0,
\end{align}
where $\lambda_2$ controls the contribution of $\bm{E}_L$.

\subsection{Modal Regression based Structured Low-rank Matrix Recovery}
Inspired by \cite{wang2017cognitive, wang2017modal}, we flexibly incorporate modal regression into Eq.~(\ref{obj:struct2}) to learn a more satisfactory mapping function even when data is contaminated by complicated noise. Under the zero-mode noise assumption, we have:
\begin{align}
    f^* = f_M = \argmax_{f\in\mathcal{M}}R\left(f\right) = \argmin_{f\in\mathcal{M}}\mathcal{L}_M\left(\bm{E}\right),
\end{align}
where $\mathcal{L}_M\left(\bm{E}\right)$ is the modal regression based loss function. According to Eq.~(\ref{prob_e}), $\mathcal{L}_M\left(\bm{E}\right)$ can be defined as:
\begin{align}\label{l_m}
    \mathcal{L}_M\left(\bm{E}\right) = dm\left(1-p_{\bm{E}}\left(0\right)\right),
\end{align}
Note that Parzen window method~\cite{parzen1962estimation} is commonly used to estimate an unknown probability density~\cite{silverman2018density,scott2015multivariate}:
\begin{align}\label{estimation_p_e}
    p_{\bm{E}}\left(0\right) = \frac{1}{dm}\sum_{i=1}^d\sum_{j=1}^m\kappa\left(\bm{E}_{ij}-0\right),
\end{align}
where $\kappa$ is a kernel function, such as Gaussion kernel or Epanechnikov kernel.

Combining Eqs.~(\ref{l_m}) with (\ref{estimation_p_e}), we have 
\begin{align}\label{def:LossMR}
\mathcal{L}_M\left(\bm{E}\right) = \sum_{i=1}^d\sum_{j=1}^m\left(1-\kappa\left(\bm{E}_{ij}\right)\right).
\end{align}

Incorporating $\mathcal{L}_M$ into SLMR, we have
\begin{align}\label{obj:MR-SLMR}
&\min_{\bm{Z},\bm{E},\bm{E}_L,\bm{P}}\norm{\bm{Z}}_{*}\!+\!\lambda_{1}\mathcal{L}_M\left(\bm{E}\right)+\lambda_{2}\norm{\bm{E}_L}_F^2\nonumber\\
&s.t.\quad \bm{P}^\mathrm{T}\bm{X}\!=\!\bm{P}^\mathrm{T}\bm{X}\bm{Z}+\bm{E},~\bm{L}\bm{Z}\!=\!\bm{L}+\bm{E}_L,~\bm{Z}\!\geq\!0.
\end{align}
Eq.~(\ref{obj:MR-SLMR}) is the proposed MR-SLMR.

\subsection{Discussion on MR-SLMR}
In this part, we discuss the error term $\bm{E}$ in Eq.~(\ref{obj:MR-SLMR}) in detail. From the in-depth analysis, we can find that $\bm{E}$ is actually used to fit the noise of projected samples rather than original samples. Moreover, it is worth noting that complicated noise is generally applied to original samples. In this case, noise models seem to sometimes fail to work even if noise obeys predefined distributions. However, we still utilize modal regression to model error from projected samples in this paper even when the error sometimes does not obey the zero-mode distributions. This is because we expect that MR-SLMR can minimize the role of features corrupted by noise as much as possible when we learn a mapping function. This is also the reason why other algorithms use the same scheme. Experimental results in Sect.~\ref{subsec:face_recognition} indicate the robustness of our MR-SLMR to various complicated noise.

\subsection{Solution to MR-SLMR}
\label{subsec:optimization}
Problem~(\ref{obj:MR-SLMR}) can be well solved by Augmented Lagrangian Multiplier (ALM)~\cite{xu2016discriminative}. We first introduce a relax variable $\bm{J}$, then Eq.~(\ref{obj:MR-SLMR}) can be translated into
\begin{align}\label{relax_variable}
\min_{\bm{J},\bm{Z},\bm{E},\bm{E}_L,\bm{P}}&\norm{\bm{J}}_{*}\!+\!\lambda_{1}\mathcal{L}_M\left(\bm{E}\right)\!+\!\lambda_{2}\norm{\bm{E}_L}_F^2\nonumber\\
s.t.\quad &\bm{P}^\mathrm{T}\bm{X}\!=\!\bm{P}^\mathrm{T}\bm{X}\bm{Z}+\bm{E},\quad \bm{J}\!=\!\bm{Z},\nonumber\\
&\bm{L}\bm{Z}\!=\!\bm{L}\!+\!\bm{E}_L,\quad \bm{Z}\!\geq\!0,
\end{align}
where the augmented Lagrangian function is formulated as
\begin{align}\label{alf}
\norm{\bm{J}}_{*}&\!+\!\lambda_1\mathcal{L}_M\left(\bm{E}\right)\!+\!\lambda_2\norm{\bm{E}_L}_F^2\nonumber\\ &\!+\!\mathrm{tr}\left(\bm{Y}_1^\mathrm{T}\left(\bm{P}^\mathrm{T}\bm{X}\!-\!\bm{P}^\mathrm{T}\bm{X}\bm{Z}\!-\!\bm{E}\right)\right)\!+\!\mathrm{tr}\left(\bm{Y}_2^\mathrm{T}\left(\bm{Z}\!-\!\bm{J}\right)\right)\nonumber\\
&\!+\!\mathrm{tr}\left(\bm{Y}_3^\mathrm{T}\left(\bm{L}\bm{Z}\!-\!\bm{L}\!-\!\bm{E}_L\right)\right)\!+\!\frac{\mu}{2}\norm{\bm{L}\bm{Z}\!-\!\bm{L}\!-\!\bm{E}_L}_F^2\nonumber\\
&\!+\!\frac{\mu}{2}\left(\norm{\bm{P}^\mathrm{T}\bm{X}\!-\!\bm{P}^\mathrm{T}\bm{X}\bm{Z}\!-\!\bm{E}}_F^2\!+\!\norm{\bm{Z}\!-\!\bm{J}}_F^2\right),
\end{align}
where $\bm{Y}_1$, $\bm{Y}_2$, and $\bm{Y}_3$ are Lagrange multipliers, and $\mu>0$ is a penalty parameter. Here, alternating minimization is used to optimize $\bm{J}$, $\bm{Z}$, $\bm{E}$, $\bm{E}_L$ and $\bm{P}$. Specifically, when we optimize one variable in the $\left(t\!+\!1\right)$-th iteration, others are set as their latest available values. To facilitate understanding, we define $\bm{J}_t$ as variable $\bm{J}$ in the $t$-th iteration~(similar expression for others), and above variables can be updated in the $\left(t\!+\!1\right)$-th iteration as follows:

\noindent \textbf{Updating $\bm{J}$}:
\begin{align}\label{slove_J}
\bm{J}_{t+1} \!=\! \arg\min_{\bm{J}}\frac{1}{\mu_t}\norm{\bm{J}}_*\!+\!\frac{1}{2}\norm{\bm{J}\!-\!\left(\bm{Z}_t\!+\!\frac{\bm{Y}_{2,t}}{\mu_t}\right)}_F^2.
\end{align}
Eq.~(\ref{slove_J}) can be optimized by the singular value thresholding~(SVT)~\cite{lin2010augmented}.

\noindent \textbf{Updating $\bm{E}$}:
\begin{align}\label{solve_E_modal}
\bm{E}_{t+1} \!=\! \arg\min_{\bm{E}}&\lambda_1\mathcal{L}_M\left(\bm{E}\right)\!+\!\mathrm{tr}\left(\bm{Y}_{1,t}^\mathrm{T}\left(\bm{P}_t^\mathrm{T}\bm{X}\!-\!\bm{P}_t^\mathrm{T}\bm{X}\bm{Z}_t\!-\!\bm{E}\right)\right)\nonumber\\
&\!+\!\frac{\mu_t}{2}\norm{\bm{P}_t^\mathrm{T}\bm{X}\!-\!\bm{P}_t^\mathrm{T}\bm{X}\bm{Z}_t\!-\!\bm{E}}_F^2.
\end{align}

\noindent Problem~(\ref{solve_E_modal}) can be optimized by the half-quadratic~(HQ) theory~\cite{nikolova2005analysis}. For the function $\phi\left(u\right)=1-\kappa\left(u\right)$, we have 
\begin{align}\label{Half_quadratic}
\phi\left(u\right)\!=\!\inf\left(\frac{1}{2}vu^2\!+\!\psi\left(v\right)\right),
\end{align}
where the function $\psi\left(v\right)$ is the dual convex function of $\phi\left(u\right)$~\cite{nikolova2005analysis}, $v\!\in\!\mathbb{R}$ and $u\!\in\!\mathbb{R}$. The infimum of Eq.~(\ref{Half_quadratic}) is reached at 
\begin{align}\label{infimum_HQ}
v\!=\!\tau\left(u\right)\!\coloneqq\!
\begin{cases}
\phi^{''}\left(0^+\right),\\
\phi^{'}\left(u\right)/u,
\end{cases}\!=\!
\begin{cases}
-\kappa^{''}\left(0^+\right),&u=0\\
-\kappa^{'}\left(u\right)/u,&u\neq0.
\end{cases}
\end{align}
Hence, for the Gaussian kernel $\kappa\left(u\right) = \frac{1}{\sqrt{2\pi}\sigma}\exp\left(-u^2/2\sigma^2\right)$, we have $v\!=\!\tau\left(u\right)\!=\!\frac{1}{\sigma^2}\kappa\left(u\right)$. Using Eq.~(\ref{Half_quadratic}), problem~(\ref{def:LossMR}) can be reformulated as
\begin{align}\label{modal_E_HQ}
\mathcal{L}_M\left(\bm{E}\right)&=\min\sum_{i=1}^p\sum_{j=1}^{m}\frac{1}{2}\bm{W}_{ij}\bm{E}_{ij}^2\!+\!\sum_{i=1}^p\sum_{j=1}^{m}\psi\left(\bm{W}_{ij}\right)\nonumber\\
&\!=\!\min\frac{1}{2}\norm{\bm{W}^{\frac{1}{2}}\circ{\bm{E}}}_F^2\!+\!\sum_{i=1}^p\sum_{j=1}^{m}\psi\left(\bm{W}_{ij}\right),
\end{align}
where $\bm{W}\in\mathbb{R}^{p\times{m}}$ and $\circ$ denotes the Hadamard product. Thus, problem~(\ref{solve_E_modal}) can be rewritten as 
\begin{align}\label{solve_E_modal_HQ}
&\bm{E}_{t+1} \!=\! \arg\min_{\bm{E}}\frac{\lambda_1}{2}\norm{\bm{W}^{\frac{1}{2}}\circ{\bm{E}}}_F^2\!+\!\lambda_1\sum_{i=1}^p\sum_{j=1}^{m}\psi\left(\bm{W}_{ij}\right)\nonumber\\
&\!+\!\frac{\mu_t}{2}\norm{\bm{P}_t^\mathrm{T}\bm{X}\!-\!\bm{P}_t^\mathrm{T}\bm{X}\bm{Z}_t\!-\!\bm{E}}_F^2\!+\!\mathrm{tr}\left(\bm{Y}_{1,t}^\mathrm{T}\left(\bm{P}_t^\mathrm{T}\bm{X}\!-\!\bm{P}_t^\mathrm{T}\bm{X}\bm{Z}_t\!-\!\bm{E}\right)\right).
\end{align}
According to the HQ theory~\cite{nikolova2005analysis}, the noise term $\bm{E}$ in problem~(\ref{solve_E_modal_HQ}) can be obtained by the following alternate procedure
\begin{align}\label{update_W}
\bm{W}_{ij}\!=\!\tau\left(\bm{E}_{ij}\right),\{i=1,\cdots,p;j=1,\cdots,m\},
\end{align}
\begin{align}\label{update_E}
\bm{E} \!=\! \arg\min_{\bm{E}}&\frac{\lambda_1}{2}\norm{\bm{W}^{\frac{1}{2}}\circ{\bm{E}}}_F^2\!+\!\mathrm{tr}\left(\bm{Y}_{1,t}^\mathrm{T}\left(\bm{P}_t^\mathrm{T}\bm{X}\!-\!\bm{P}_t^\mathrm{T}\bm{X}\bm{Z}_t\!-\!\bm{E}\right)\right)\nonumber\\
&\!+\!\frac{\mu_t}{2}\norm{\bm{P}_t^\mathrm{T}\bm{X}\!-\!\bm{P}_t^\mathrm{T}\bm{X}\bm{Z}_t\!-\!\bm{E}}_F^2.
\end{align}
If the Gaussian kernel is adopted, the scale parameter $\sigma$ is empirically determined by $\sigma=\left(\frac{1}{2pm}\norm{\bm{E}}_F^2\right)^{\frac{1}{2}}$~\cite{principe2000information}. Then problem~(\ref{update_W}) can be solved by Eq.~(\ref{infimum_HQ}), whereas problem~(\ref{update_E}) can be tackled by 
\begin{align}\label{close_form_E}
    \bm{E}\!=\!\left(\bm{P}_t^\mathrm{T}\left(\bm{X}-\bm{X}\bm{Z}_t\right)+\frac{\bm{Y}_{1,t}}{\mu_t}\right)./\left(\frac{\lambda_1}{\mu_t}\bm{W}+\bm{D}\right),
\end{align}
where $./$ represents entrywise division and $\bm{D}\!\in\!\mathbb{R}^{p\times{m}}$ is a matrix with all elements equal to one. The HQ theory guarantees that the iterations above converge.

\noindent \textbf{Updating $\bm{E}_L$}:
\begin{align}\label{slove_E_L}
\bm{E}_{L,t+1} \!=\! (2\lambda_2\!+\!\mu_t)^{-1}\left(\bm{Y}_{3,t}\!+\!\mu_{t}\bm{L}\bm{Z}_t\!-\!\mu_t\bm{L}\right).
\end{align}

\noindent \textbf{Updating $\bm{Z}$}:
\begin{align}\label{slove_Z}
\bm{Z} \!=\! \bm{Z}_1^{-1}\bm{Z}_2,
\end{align}
where $\bm{Z}_1$ and $\bm{Z}_2$ are represented as follows:
\begin{align}
&\bm{Z}_1 \!=\! \bm{X}^\mathrm{T}\bm{P}_{t}\bm{P}_t^\mathrm{T}\bm{X}\!+\!\bm{\mathrm{I}}+\bm{L}^\mathrm{T}\bm{L},\!\nonumber\\
&\bm{Z}_2\!=\!\bm{X}^\mathrm{T}\bm{P}_{t}\left(\bm{P}_t^\mathrm{T}\bm{X}\!-\!\bm{E}_t\right)\!+\!\bm{J}_t\!+\!\bm{L}^\mathrm{T}\left(\bm{L}\!+\!\bm{E}_{L,t}\right)\nonumber\\
&\quad\quad\quad\!+\!\left(\bm{X}^\mathrm{T}\bm{P}_t\bm{Y}_{1,t}\!-\!\bm{Y}_{2,t}\!-\!\bm{L}^\mathrm{T}\bm{Y}_{3,t}\right)/\mu_{t}.\nonumber
\end{align}

\noindent \textbf{Updating $\bm{P}$}:
\begin{align}\label{slove_P}
\bm{P}_{t+1} = &\left(\left(\bm{X}\!-\!\bm{X}\bm{Z}_t\right)\left(\bm{X}\!-\!\bm{X}\bm{Z}_t\right)^\mathrm{T}\right)^{-1}\nonumber\\
&\quad\quad\left(\left(\bm{X}\!-\!\bm{X}\bm{Z}_t\right)\left(\bm{E}_t^\mathrm{T}\!-\!\bm{Y}_{1,t}^\mathrm{T}/\mu_{t}\right)\right).
\end{align}

Afterwards, we update multipliers $\bm{Y}_1$, $\bm{Y}_2$, and $\bm{Y}_3$ in the following way
\begin{align}\label{multipliers_update}
&\bm{Y}_{1,t+1} \!=\! \bm{Y}_{1,t}\!+\!\mu_t\left(\bm{P}_{t+1}^\mathrm{T}\bm{X}\!-\!\bm{P}_{t+1}^\mathrm{T}\bm{X}\bm{Z}_{t+1}\!-\!\bm{E}_{t+1}\right),\nonumber\\
&\bm{Y}_{2,t+1} \!=\! \bm{Y}_{2,t}\!+\!\mu_t\left(\bm{Z}_{t+1}\!-\!\bm{J}_{t+1}\right),\nonumber\\
&\bm{Y}_{3,t+1} \!=\! \bm{Y}_{3,t}\!+\!\mu_t\left(\bm{L}\bm{Z}_{t+1}\!-\!\bm{L}\!-\!\bm{E}_{L,t+1}\right),\nonumber\\
&\mu_{t+1}\!=\!\min\left(\rho\mu_t,\mu_{max}\right),
\end{align}
where $\rho > 1$ and $\mu_{max}$ is a constant. The detailed iteration process is summarized in Algorithm 1.
%~\uppercase\expandafter{\romannumeral1}.

\begin{algorithm2e}
\caption{Solving Problem (\ref{obj:MR-SLMR}) by ADMM}
\SetKwInput{Init}{Initialization}
\SetKw{return}{Return}
\DontPrintSemicolon
\KwIn{$\bm{X}$, $\bm{L}$, $\lambda_{1}$ and $\lambda_{2}$;}
\KwOut{$\bm{P}$;}
\Init{$\bm{J} = \bm{Z} = \bm{E} = \bm{E}_L = \bm{P} = 0$,\
$\bm{Y}_1 = \bm{Y}_2 = \bm{Y}_3 = 0$,\quad\quad\quad\
$\mu = 10^{-3}$,\quad
$\mu_{max} = 10^6$,\quad$t_{max}=10^3$,
$\rho = 1.03$,\quad$\epsilon = 10^{-6}$;\;}
\While{$t\leq{t_{max}}$}
{
   1.~Update $\bm{J}$ by solving problem~(\ref{slove_J}); \;
   2.~Update $\bm{E}$ by solving problem~(\ref{solve_E_modal});\;
   3.~Update $\bm{E}_L$ by~(\ref{slove_E_L});\;
   4.~Update $\bm{Z}$ by~(\ref{slove_Z}), and then $\bm{Z}=\max{\left(0, \bm{Z}\right)}$~\cite{yin2015dual};\;
   5.~Update $\bm{P}$ by~(\ref{slove_P});\;
   6.~Update multipliers and parameter $\mu$ by (\ref{multipliers_update});\;
   \If{the following conditions meet \;
   \quad\quad$\norm{\bm{P}^\mathrm{T}\bm{X}\!-\!\bm{P}^\mathrm{T}\bm{XZ}\!-\!\bm{E}}_\infty < \epsilon$,\;
   \quad\quad$\norm{\bm{Z}\!-\!\bm{J}}_\infty < \epsilon$,\;
   \quad\quad$\norm{\bm{L}\bm{Z}\!-\!\bm{L}\!-\!\bm{E}_L}_\infty < \epsilon$,\;
   }{break;\;}
   $t\!=\!t+1$;\;
}
\return{$\bm{P},\bm{Z},\bm{E},\bm{E}_L$;\;}
\end{algorithm2e}

\subsection{Complexity Analysis}
For simplicity, we first focus on the analysis of the computational cost for MR-SLMR in the $t$-th iteration. According to our optimization for Eq.~(\ref{slove_J}), the main computational cost is from the SVD of $\left(\!\bm{Z}_t\!+\!\bm{Y}_{2,t}/\mu_t\!\right)$. Hence, the computational complexity of $\bm{J}$ is close to $\mathcal{O}\!\left(\!m^3\!\right)$. Moreover, $\bm{E}$ is updated by alternately optimizing Eqs.~(\ref{update_W}) and (\ref{update_E}). Assume that the optimization procedure is converged in the $T$-th iteration, the computational complexity of $\bm{E}$ is approximately $\mathcal{O}\!\left(\!T\left(\!m^2d\!+\!mpd\!\right)\!\right)$. Finally, we can easily derive the computational complexity of $\bm{E}_L$, $\bm{Z}$ and $\bm{P}$ from Eqs.~(\ref{slove_E_L}), (\ref{slove_Z}) and (\ref{slove_P}), where the computational complexity of $\bm{E}_L$  is close to $\mathcal{O}\!\left(\!Cm^2\!\right)$, while the cost of $\bm{Z}$ and $\bm{P}$ is approximately $\mathcal{O}\!\left(\!mpd\!+\!m\!\left(\!p\!+\!C\!\right)\!+\!m^3\!\right)$ and $\mathcal{O}\!\left(\!m^2d\!+\!mpd\!+\!d^2\left(d\!+\!p\!+\!m\!\right)
\!\right)$, respectively. Therefore, the computational complexity of MR-SLMR in the $t$-th iteration is close to $\mathcal{O}\!\left(\!m^3\!+\!T\!\left(\!m^2d\!+\!mpd\!\right)\!+\!m^2C\!+\!d^2\!\left(d\!+\!p\!+\!m\!\right)\!\right)$. To summarize, the computational complexity of MR-SLMR is about $\mathcal{O}\!\left(\!T_1\left(\!m^3\!+\!T\!\left(\!m^2d\!+\!mpd\!\right)\!+\!m^2C\!+\!d^2\!\left(d\!+\!p\!+\!m\!\right)\!\right)\!\right)$, where $T_1$ is the iterations of MR-SLMR. 

By contrast, the computational cost of LRCS and LRDE are close to $\mathcal{O}\!\left(\!T_2\!\left(\!m^3\!+\!m^2\!\left(\!p\!+\!d\!\right)\!+\!m\!\left(\!pd\!+\!d^2\!\right)\!+\!d^3\!+\!dp\!\left(\!p\!+\!d\!\right)\!\right)\!\right)$ and $\mathcal{O}\!\left(\!T_3\!\left(\!m^3\!+\!m^2p\!+\!mpd\!+\!dp^2\!\right)\!\right)$ respectively, where $T_2$ and $T_3$ are the iterations of LRCS and LRDE. In practice, we have $p\!<\!d\!<\!m$ and $C\!<\!m$, thus the complexity of LRCS, LRDE and our MR-SLMR can be further simplified with $\mathcal{O}\!\left(\!T_2\!\left(\!m^3\!\right)\!\right)$, $\mathcal{O}\!\left(\!T_3\!\left(\!m^3\!\right)\!\right)$ and $\mathcal{O}\!\left(\!T_1\!\left(\!m^3\!+\!Tdm^2\!\right)\!\right)$. Since $Td\!<\!m$ and $T_1\!\approx\!T_2\!<\!T_3$ in most cases, three methods generally have similar computational complexity. Moreover, experimental results in Sect.~\ref{subsec:object_classification} corroborate our complexity analysis and LRDE presents a slight disadvantage in time cost due to relatively large iterations.

%In comparison, the computational complexity of RMSL and LRDE are approximately $\emph{O}\left({n}^3+n^2d+n(pd+d^2)+d^3\right)$ and $\emph{O}\left({n}^3+n^2p+npd+dp^2\right)$, respectively. In practice, we have $d<n, c<n, p\ll{d}$ and $\xi$ is at most a constant term, thus, the computational complexity of all three algorithm can be simplified with $\emph{O}\left(n^3\right)$ and the computational cost further depends on the iterations of different algorithm.
% For simplicity, we mainly discuss the computational complexity of Algorithm 1 in $t$th iteration. Note that $\bm{X}\!\in\!\mathbb{R}^{d\times{n}}$, $\bm{P}\!\in\!\mathbb{R}^{d\times{p}}$, $\bm{Z}\!\in\!\mathbb{R}^{n\times{n}}$ and $\bm{Y}\!\in\!\mathbb{R}^{c\times{n}}$ . The time consumption of Algorithm 1 mainly involves three operations, singular value decomposition (SVD), matrix inverse and matrix multiplication. 

% For all steps in Algorithm 1, problem in Step 1 is addressed by singular value decomposition. According to the dimension of relevant matrices, corresponding time consumption are $\emph{O}\left({n}^3\right)$. For step 2, an alternate optimization is used to update $\bm{E}$, Then, at each iteration, the main computational cost of update $\bm{E}$ is the matrix multiplication and the complexity is $\emph{O}\left(pn^2\right)$. The major operations in Step 3, Step 4 and Step 5 are matrix inverse and matrix multiplication. Hence, it takes $\emph{O}\left(cn^2\right)$, $\emph{O}\left(dn^2\!+\!nd^2\!+\!d^3\!+\!dnp\!+\!d^2p\right)$ and $2\emph{O}\left(n^3\right)$, respectively.
\section{Experiments}
\label{sec:experiments}
In this section, four benchmark databases for cross-view classification, namely the MNIST and the USPS handwritten digit databases\footnote{https://cs.nyu.edu/~roweis/data.html}, the Amsterdam Library of Object Images~(ALOI)\footnote{http://aloi.science.uva.nl/?tdsourcetag=s\_pctim\_aiomsg}, the Columbia University Image Library (COIL-100)\footnote{http://www.cs.columbia.edu/CAVE/software/softlib/coil-100.php} and the CMU Pose, Illumination, and Expression (PIE) database (CMU PIE), are used to evaluate the performance of LMvSL based methods. This section is arranged as follows. Sect.~\ref{subsec:evaluation_protocol} introduces evaluation manners and experimental setting. Experiment to evaluate algorithm performance on handwritten digit classification across styles is conducted in Sect.~\ref{subsec:handwritten_digits}. Furthermore, we demonstrate the efficacy of MR-SLMR on object classification across pose on the ALOI and COIL-100 datasets in Sect.~\ref{subsec:object_classification}, and we also validate its superiority and robustness on face classification across pose on the CMU PIE dataset in Sect.~\ref{subsec:face_recognition}. Finally, we evaluate the parameter sensitivity and convergence in Sect.~\ref{subsec:property_analysis}.

\begin{table*}[htb]
  \centering
  \caption{Comparison results~(\%) of all methods on handwritten digit datasets. \textbf{Bold} denotes the best result.}
    \begin{tabular}{c|c|cccccc}
    \hline
    Gallery & Probe & SRRS & LRCS & RMSL & LRDE &  CLRS &  MR-SLMR \\
    \hline
    \multirow{2}{*}{mnist} & mnist & 81.9  & 80.7  & 82.0 & 81.4  & \textbf{82.6} & 76.8 \\
                           & usps & 32.1  & 31.5  & 32.5  & 45.4  & 32.2 & \textbf{64.5} \\
    \hline
    \multirow{2}{*}{usps} & usps & \textbf{83.3}  & 82.9  & 83.2 & 73.7  & 81.9  & 74.2 \\
                          & mnist & 16.0  & 17.0  & 16.0  & 43.4  & 16.0 & \textbf{61.4} \\
    \hline
    \multicolumn{2}{c|}{Average} & 53.3$\pm$0.9  & 53.1$\pm$1.0  & 53.4$\pm$0.9  & 61.0$\pm$1.4  & 53.2$\pm$1.1   & \textbf{69.2}$\pm$1.0 \\
    \hline
    \end{tabular}
  \label{tab:handwriting}
\end{table*}

\subsection{Evaluation Protocol and Experimental Setting}
\label{subsec:evaluation_protocol}
Two evaluation protocols have been developed to evaluate algorithm performance for cross-view classification including pairwise manner and multi-view manner~\cite{kan2012multi}. The projected samples~($\bm{P}^\mathrm{T}\bm{X}$) are used for classification for all LMvSL based algorithms, and same as~\cite{kan2016multi,xu2018multi,YOU201937}, pairwise manner is employed to evaluate the algorithm performance during the testing phase. Formally speaking, given a gallery set $D_v$ in view $v$ and a probe set $D_u$ in view $u$, pairwise manner intends to predict the label of $\bm{x}\!\in\!{D_u}$ using a 1-NN classifier from $D_v$ and then reports the accuracy with respect to $D_v$ and $D_u$.
%\begin{align}
%    \rm{accuracy}\left(u, v\right) = \frac{\#\{\bm{x}:\bm{x}\in{D_u}\wedge\bar{y}=y\}}{\#\{\bm{x}:\bm{x}\in{D_u}\}},
%\end{align}
%where $\bar{y}$ and $y$ are the predicted label and true label of $\bm{x}$ respectively. 
Then, $k\times k$ results can be obtained by traversing $u$ and $v$ from 1 to $k$ at an interval of 1, and the average of results is reported as the mean accuracy~(mACC). 
%In contrast, multi-view manner aims to predict the label of $\bm{x}\!\in\!{D_u}$ using a 1-NN classifier from $\{D_1,\ldots,D_k\}$. Similarly, the average of $k$ results is reported as mACC. %samples from all views are regarded as gallery set while samples from one view serve as probe set when multi-view manner is adopted.
%Moreover, it is worth noting that we also evaluate the performance when gallery and probe sets are from the same view due to the existence of this case in practice. Hence, the definitions of two manners may be slightly different from those elsewhere.

In our experiments, LRCS~\cite{ding2014low}, SRRS~\cite{li2016learning}, RMSL~\cite{Ding:2016:RMS:3015812.3015987}, LRDE~\cite{Li2017Low} and CLRS~\cite{ding2018robust}, are selected for comparison, where the first one is an unsupervised approach, whereas the other four are supervised methods by considering label information under the framework of graph embedding. Experiments are repeatedly conducted 10 times, and the average of 10 results is reported as the final accuracy. Similar to~\cite{sharma2012generalized,kan2016multi,cao2018generalized}, Principal Component Analysis~(PCA) is used to perform preprocessing. Hyper-parameters of MR-SLMR and its competitors are determined via validation set. % Same as~\cite{kan2016multi,xu2018multi}, this paper employs pairwise manner to evaluate the algorithm performance during the testing phase.
%considering that a sample from one view prefers to match another sample from nearest views when multi-view manner is adopted. 

\begin{figure}[htb]
    \centering
    \includegraphics[height=3.5cm,width=7cm]{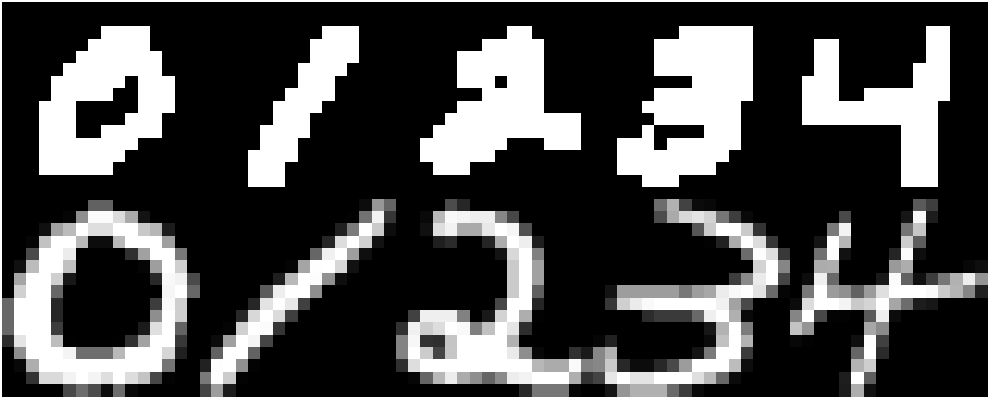}
    \caption{Exemplar subjects from  the MNIST and USPS, where subjects in the first line are from the MNIST, whereas subjects in the second line are from the USPS.}
    \label{fig:img_hw}
    \vspace{-0.5cm}
\end{figure}

\begin{figure*}[htb]
	\centering
	\begin{minipage}{3.5cm}
		\includegraphics[width=3.5cm]{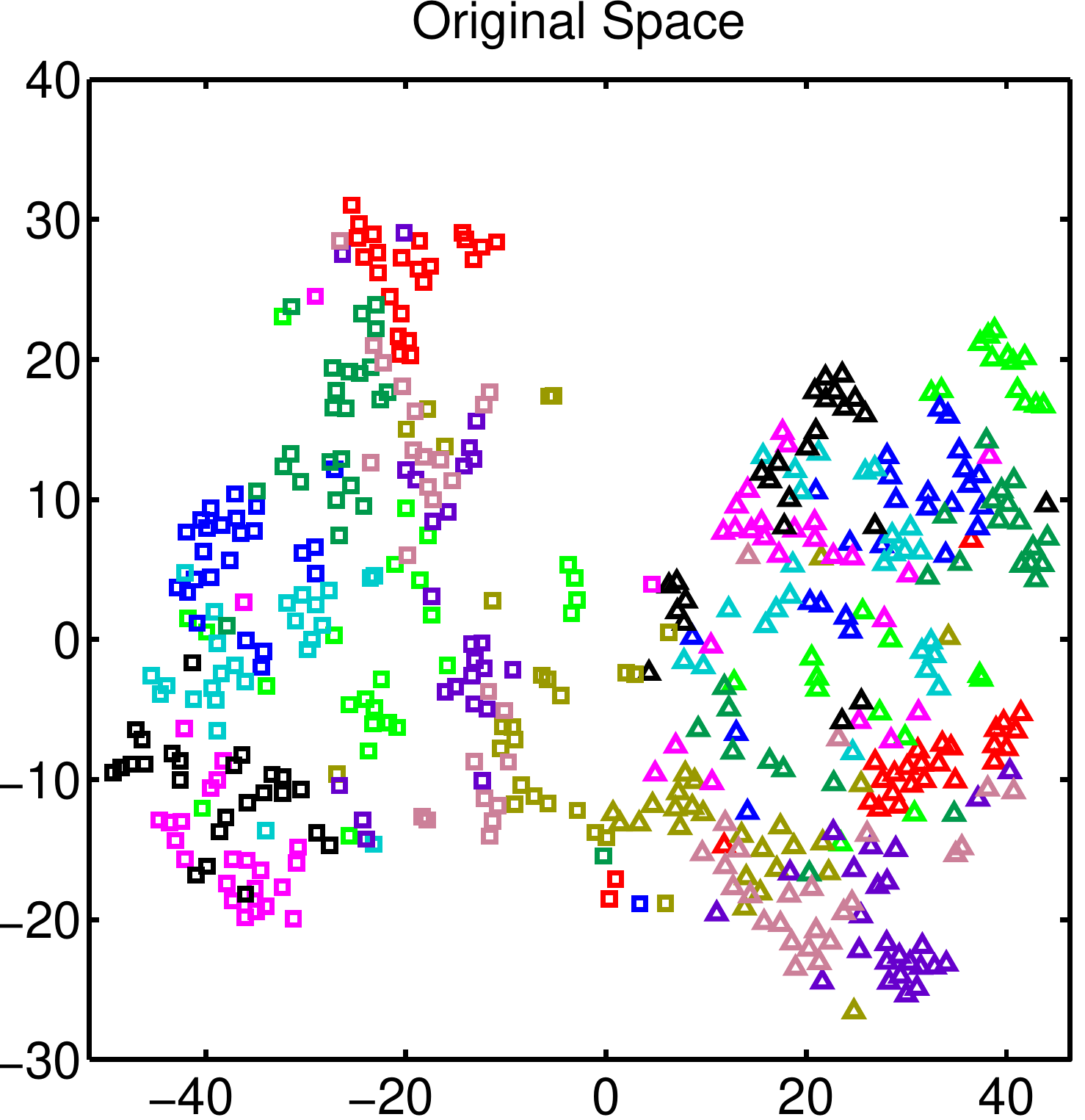}
	\end{minipage}
	\begin{minipage}{3.5cm}
		\includegraphics[width=3.5cm]{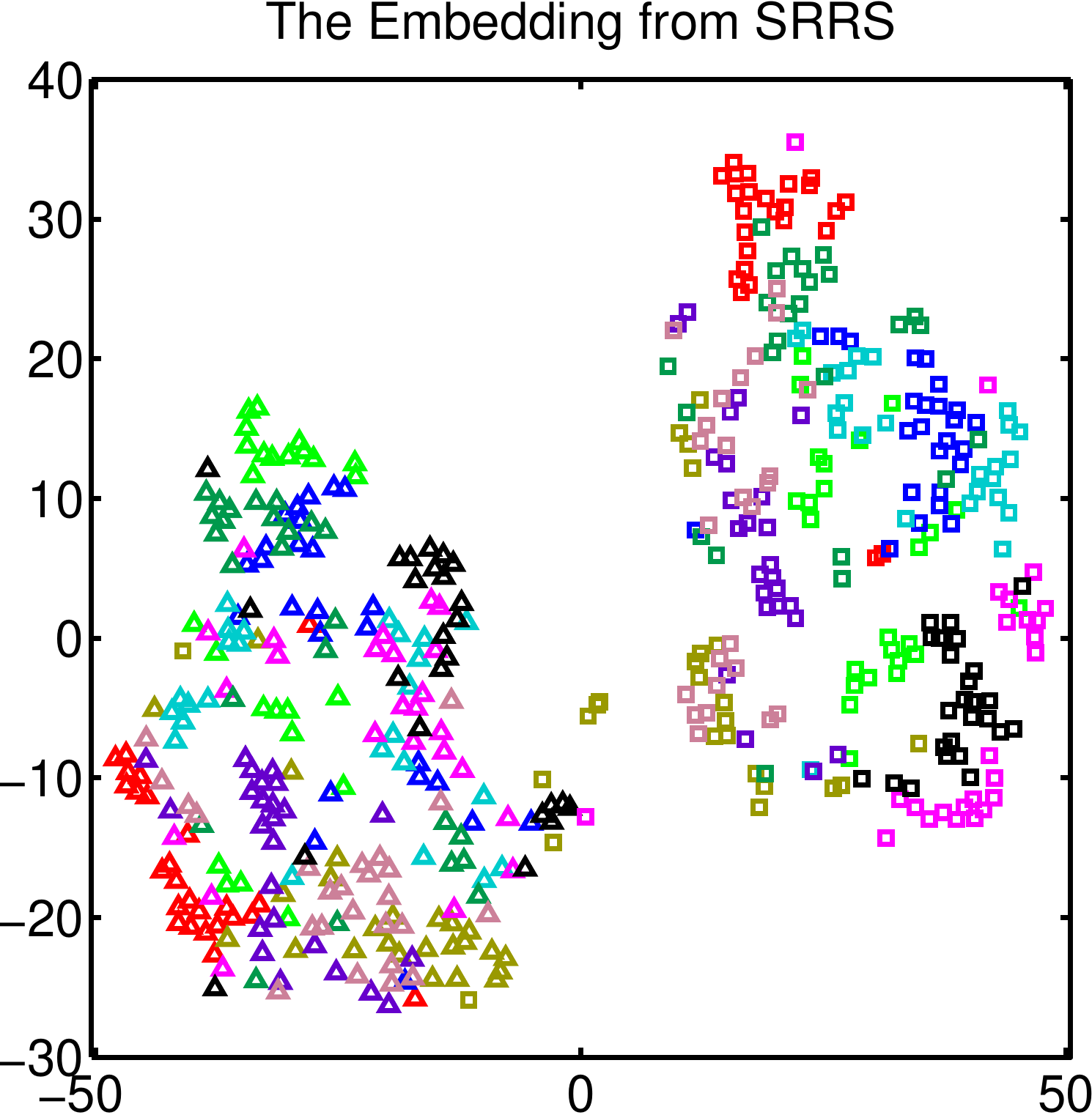}
	\end{minipage}
	\begin{minipage}{3.5cm}
		\includegraphics[width=3.5cm]{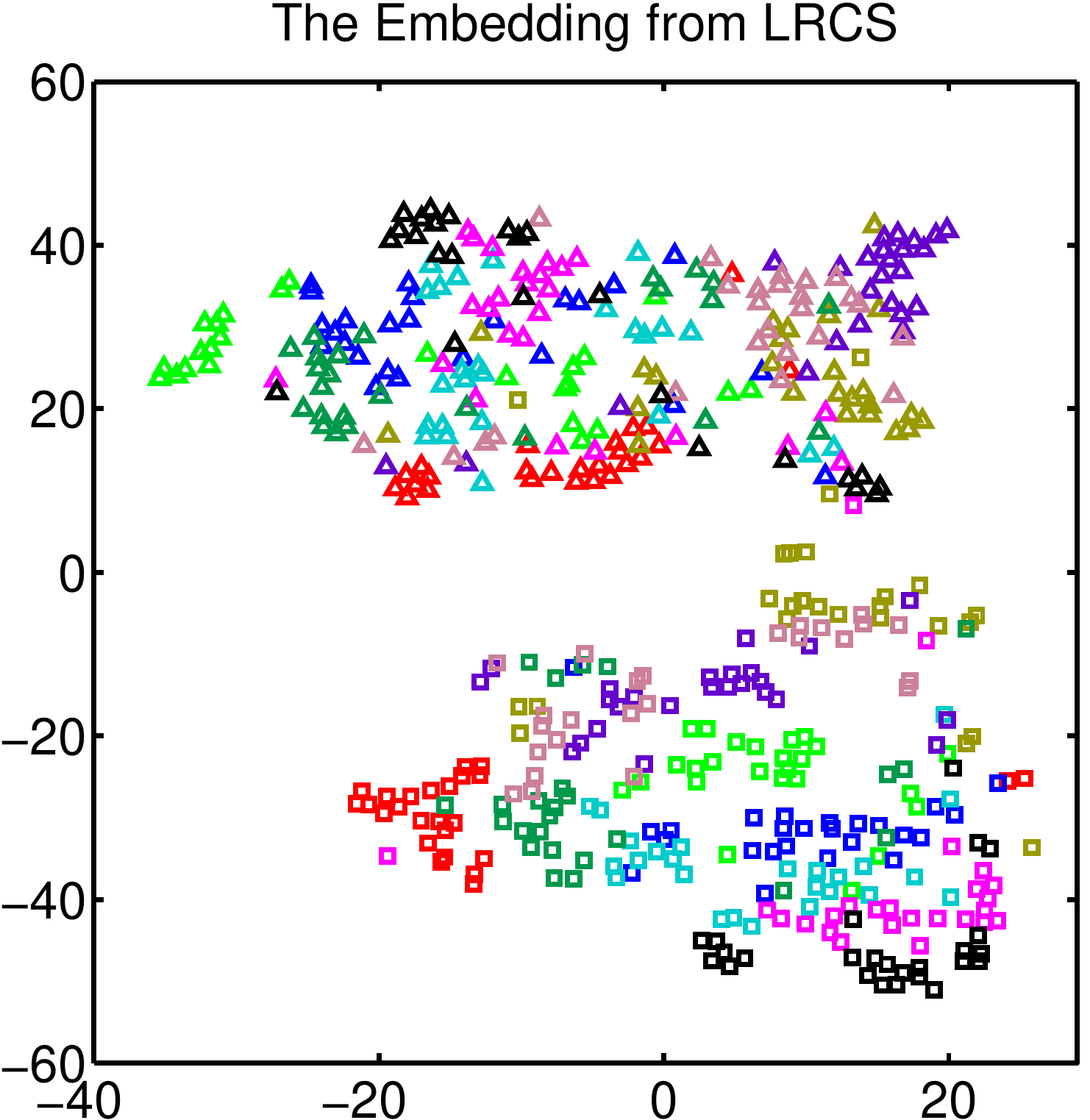}
	\end{minipage}\\
	\begin{minipage}{3.5cm}
		\includegraphics[width=3.5cm]{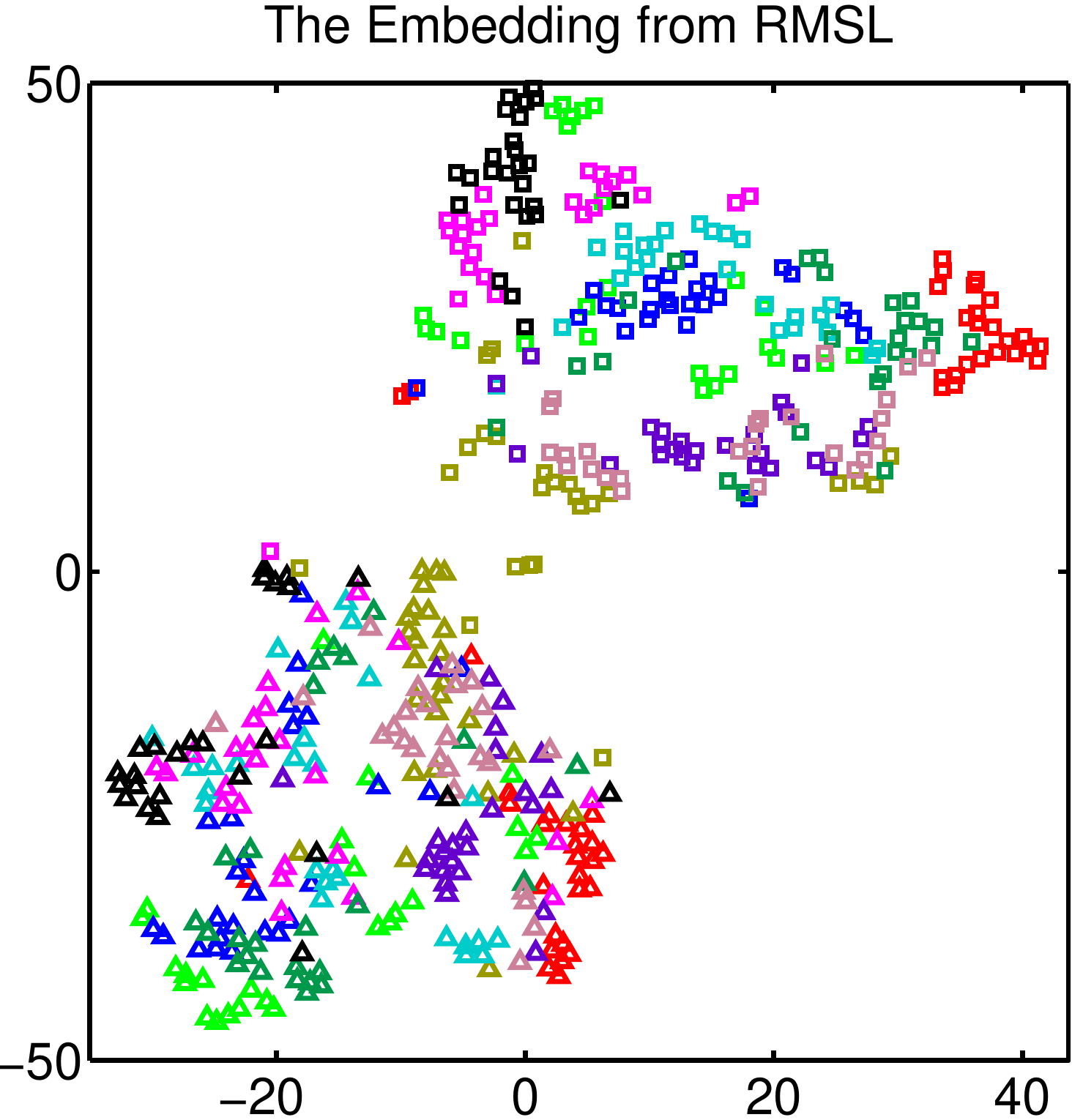}
	\end{minipage}
	\begin{minipage}{3.5cm}
		\includegraphics[width=3.5cm]{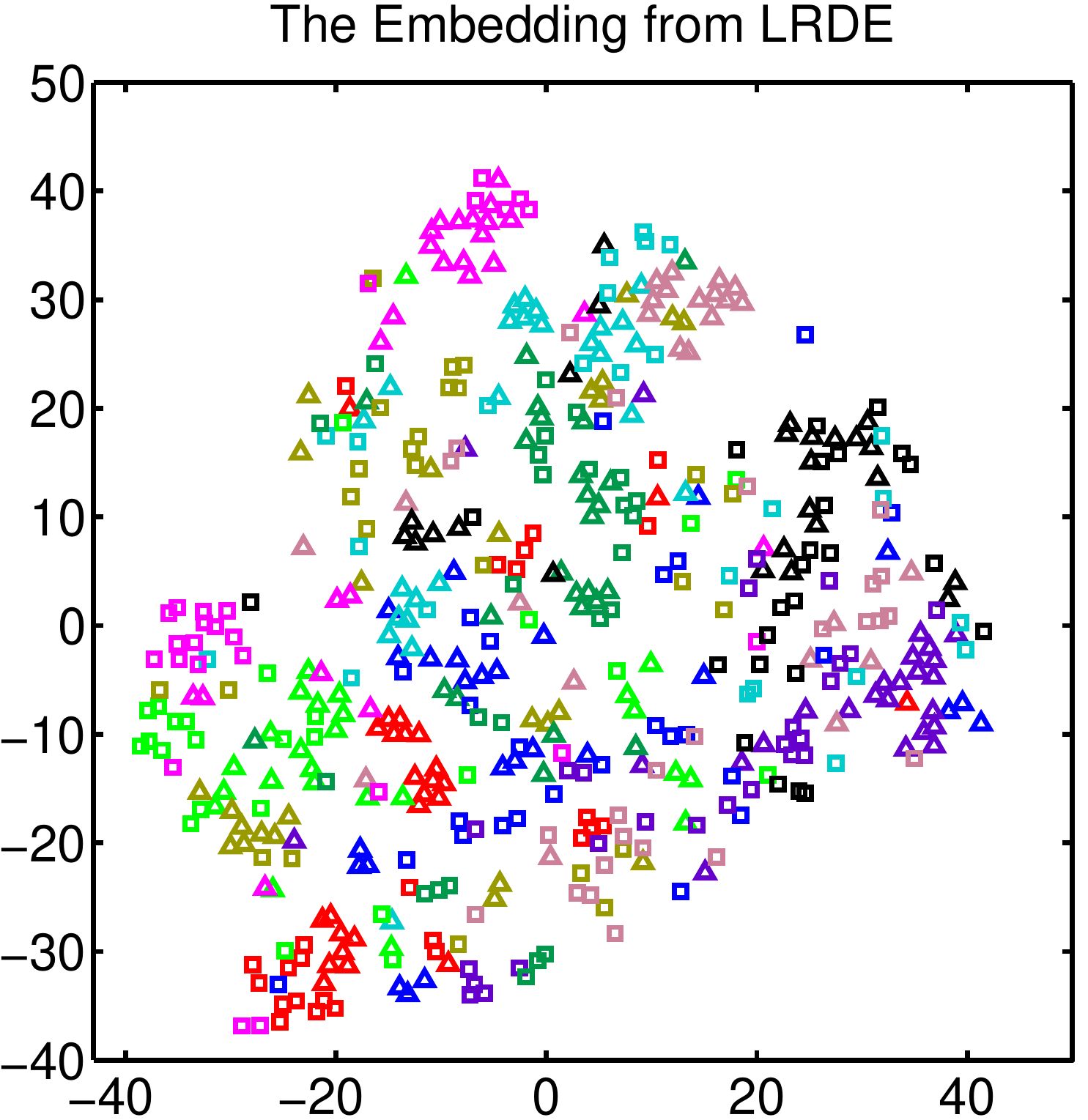}
	\end{minipage}
		\begin{minipage}{3.5cm}
		\includegraphics[width=3.5cm]{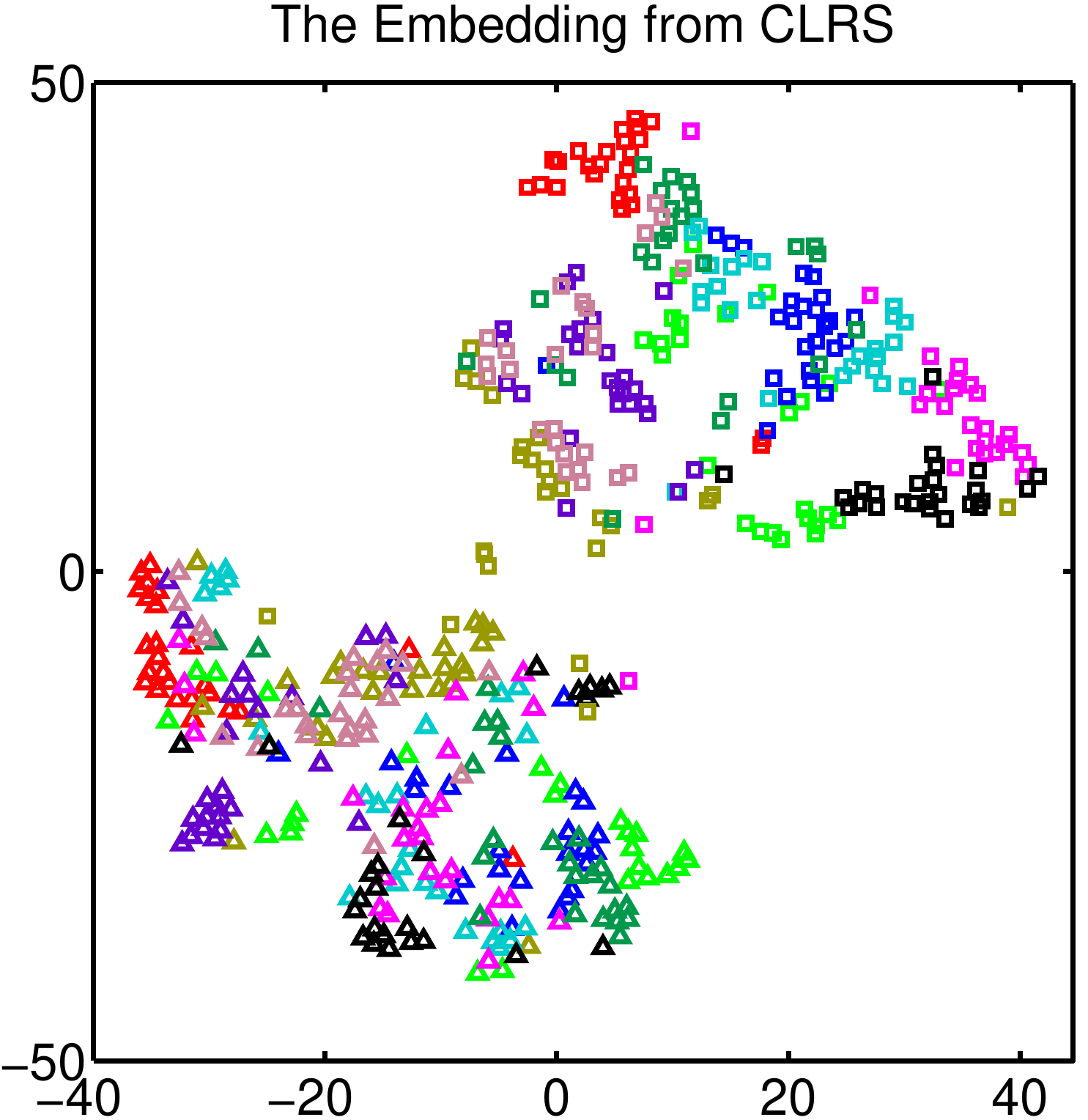}
	\end{minipage}
		\begin{minipage}{3.5cm}
		\includegraphics[width=3.5cm]{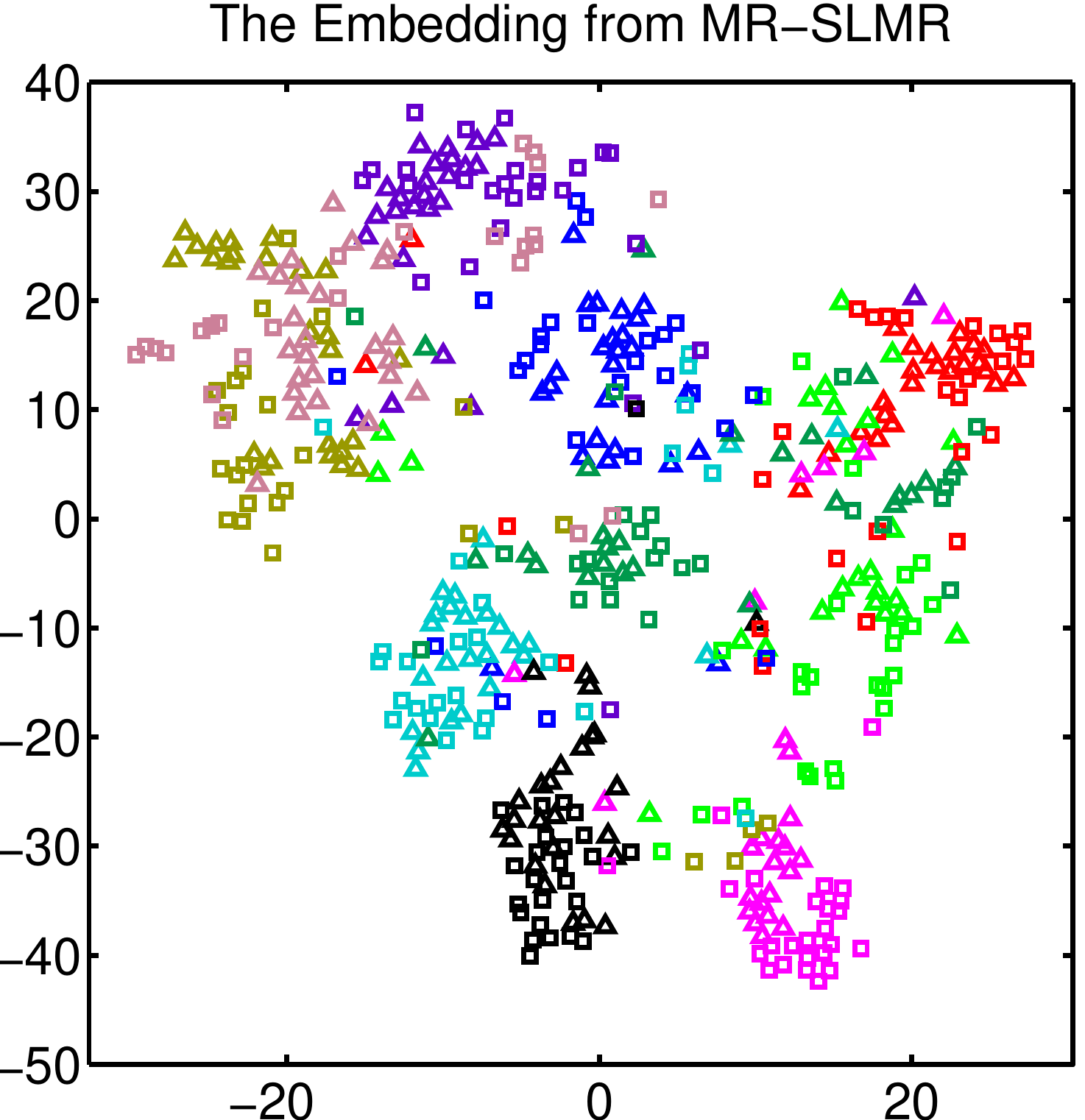}
	\end{minipage}
	\caption{Visualization of original space and the common subspace generated by LMvSL based methods on handwritten digit datasets. Markers and colors denote views and classes, respectively. \vspace{-0.2cm}}
	\label{fig:handwriting-vis}
\end{figure*}

\subsection{Handwritten Digit Classification across Styles}
\label{subsec:handwritten_digits}
Experiment to evaluate the performance of MR-SLMR on handwritten digit classification across styles is conducted on the MNIST and USPS datasets. In experiments, 1000 normalized grayscale images~(100 for each digit) of size $16\times16$ from MNIST and 1000 images~(100 for each digit) of size $16\times16$ from USPS are selected to construct multi-view dataset. Exemplar subjects are shown in Fig.~\ref{fig:img_hw}. As can be seen, it is obvious that there is a difference on handwritten style between the MNIST and USPS. For a rigorous comparison, this dataset is further divided into training set, validation set and test set at a ratio of $2\!:\!1\!:\!1$. Experimental results are summarized in Table~\ref{tab:handwriting}.

As can be seen, although SRRS, RMSL and CLRS are supervised methods, they achieve similar performance to LRCS. Moreover, these methods all perform badly when the gallery and probe data come from diverse views~(term it cross-view classification). One possible reason is that there is a large view discrepancy between MNIST and USPS, whereas above methods cannot effectively decrease it. By contrast, LRDE achieves a promising performance among competitors, since criteria A~\cite{Li2017Low} for graph embedding helps to reduce the view gap. As expected, our method achieves a remarkable improvement for cross-view classification, which can be attributed to the effective elimination of discrepancy and improvement of inter-view discriminacy by the recovery of structured low-rank matrix. However, due to the limitation of criteria B for LRDE, this method fails to provide a satisfactory result when the gallery and probe data come from the same view~(term it intra-view classification). Similarly, our MR-SLMR performs the worst for intra-view classification among competitors. This indicates that graph embedding presents advantages in enhancing intra-view discriminancy compared with structured regularization. However, experimental results on handwritten digit datasets still demonstrate the superiority of our method according to the overall performance. To make results more intuitive, subspaces of all methods are shown in Fig.~\ref{fig:handwriting-vis}. It is obvious that the embeddings corroborate the classification accuracy shown in Table~\ref{tab:handwriting}. Furthermore, one should note that within-class samples from different datasets are separated in the original space. It further indicates the large view divergence between the MNIST and the USPS.

\subsection{Object Classification across Pose}
\label{subsec:object_classification}
\begin{figure}[htb]
    \centering
    \includegraphics[height=3.6cm,width=8cm]{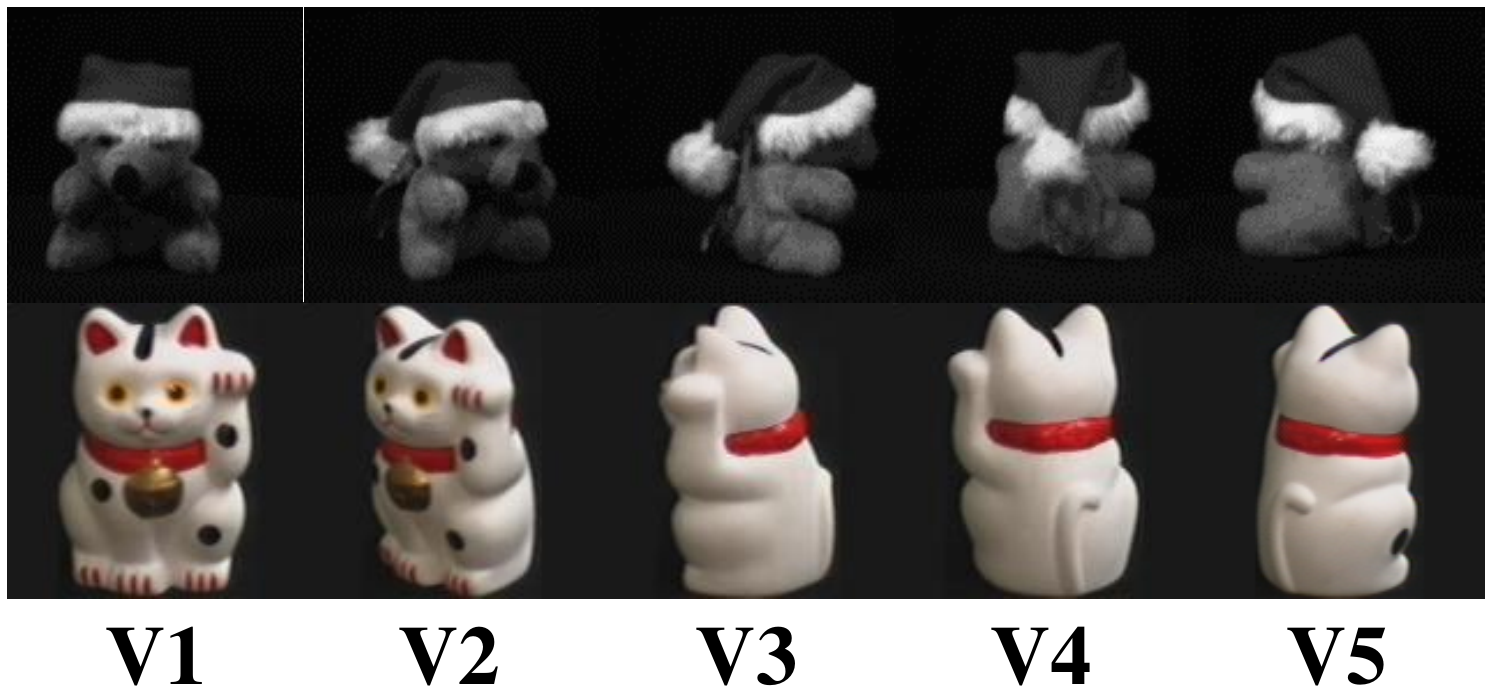}
    \caption{Exemplar subjects from ALOI and COIL-100, where subjects in the first line are from ALOI, whereas subjects in the second line are from COIL-100.}
    \label{fig:img_object}
\end{figure}

We compare MR-SLMR with other LMvSL based counterparts on the Amsterdam Library of Object Images~(ALOI) and the Columbia University Image Library~(COIL-100) datasets respectively to demonstrate the efficacy of our method. These two sets are both used to evaluate object classification across pose, where ALOI~(or COIL-100) contains images of 1000 objects~(or 100 objects), captured from $0^\circ$ to $355^\circ$ at intervals of $5^\circ$. As shown in Fig.~\ref{fig:img_object}, same as~\cite{xu2018multi}, five poses, namely V1: [$0^\circ$, $15^\circ$], V2: [$40^\circ$, $55^\circ$], V3: [$105^\circ$, $120^\circ$], V4: [$160^\circ$, $175^\circ$] and V5: [$215^\circ$, $230^\circ$] from the first 100 objects of ALOI~(or COIL-100) are selected to construct multi-view data. Similar to~\cite{ding2018robust}, experiments on AOIL~(or COIL-100) are conducted in four cases including case 1: $\{\textrm{V}1, \textrm{V}2\}$, case 2: $\{\textrm{V}1, \textrm{V}2, \textrm{V}3\}$, case 3: $\{\textrm{V}1, \textrm{V}2, \textrm{V}3, \textrm{V}4\}$ and case 4: $\{\textrm{V}1, \textrm{V}2, \textrm{V}3, \textrm{V}4, \textrm{V}5\}$. In experiments, 50 objects from ALOI~(or COIL-100) are used for training, while the remaining are divided into validate set and test set at a ratio of $1:1$. Experimental results on ALOI and COIL-100 are presented in Table~\ref{tab:ALOI} and \ref{tab:COIL}, respectively.

\begin{table}[htbp]
  \centering
  \caption{Comparison results~(\%) of all methods on ALOI object dataset.}
    \begin{tabular}{c|c|c|c|c}
    \hline
    Methods& case 1 & case 2 & case 3 & case 4 \\
    \hline
    SRRS  & 93.2$\pm$1.1 & 83.3$\pm$2.2 & 80.6$\pm$2.4 & 76.9$\pm$2.9 \\
    LRCS  & 92.4$\pm$1.2 & 81.2$\pm$1.4 & 77.3$\pm$1.6 & 72.5$\pm$1.8 \\
    RMSL  & 92.6$\pm$0.9 & 82.0$\pm$2.7 & 78.8$\pm$2.6 & 75.6$\pm$2.8 \\
    LRDE  & 93.5$\pm$1.2 & 86.7$\pm$2.2 & 82.1$\pm$2.1 & 79.5$\pm$3.0 \\
    CLRS  & 92.2$\pm$0.9 & 81.9$\pm$2.5 & 78.8$\pm$2.9 & 76.0$\pm$2.6 \\
    MR-SLMR & \textbf{94.0}$\pm$1.1 & \textbf{87.6}$\pm$1.7 & \textbf{85.3}$\pm$2.4 & \textbf{85.8}$\pm$1.6 \\
    \hline
    \end{tabular}%
    \vspace{-0.2cm}
  \label{tab:ALOI}%
\end{table}%

\begin{table}[htb]
  \centering
  \caption{Comparison results~(\%) of all methods on COIL-100 object dataset.}
    \begin{tabular}{c|c|c|c|c}
    \hline
     Methods& case 1 & case 2 & case 3 & case 4 \\
    \hline
    SRRS  & 89.3$\pm$2.0 & 75.4$\pm$2.2 & 72.8$\pm$2.1 & 73.0$\pm$2.1 \\
    LRCS  & 89.2$\pm$2.0 & 75.8$\pm$2.4 & 73.4$\pm$2.4 & 73.8$\pm$2.1 \\
    RMSL  & 89.9$\pm$2.1 & 75.8$\pm$2.4 & 73.5$\pm$2.4 & 74.0$\pm$2.5 \\
    LRDE  & 90.4$\pm$1.9 & 78.0$\pm$2.9 & 75.7$\pm$2.6 & 73.8$\pm$2.3 \\
    CLRS  & 89.2$\pm$1.9 & 75.3$\pm$2.6 & 73.0$\pm$2.8 & 73.0$\pm$2.4 \\
    MR-SLMR & \textbf{94.4}$\pm$1.6 & \textbf{84.1}$\pm$2.1 & \textbf{81.2}$\pm$2.9 & \textbf{80.9}$\pm$2.2 \\
    \hline
    \end{tabular}%
    \vspace{-0.2cm}
  \label{tab:COIL}%
\end{table}%

As shown in Table~\ref{tab:ALOI}, the performance of LRCS ranks the lowest for the lack of discriminant information. SRRS, RMSL and CLRS perform better than LRCS in almost all cases by taking into consideration of supervised information. Benefiting from the effect of criteria A on the mitigation of view discrepancy, LRDE outperforms the other four methods. Our MR-SLMR obtains a large improvement compared with LRDE. This indicates that the recovery of structured low-rank matrix did help to remove view discrepancy and improve discriminancy. Moreover, as can be seen in Table~\ref{tab:COIL}, experimental results on COIL-100 are basically consistent with those in Table~\ref{tab:ALOI}. To verify the computational complexity of algorithms, the training time in all cases of COIL-100 is reported in Table~\ref{tab:time_coil}, where experiments are conducted in Matlab R2017b with CPU i5-3470 and 8.0GB memory size. As can be seen, our MR-SLMR also presents an advantage in computational cost.

\begin{table}[htb]
  \centering
  \caption{The training time~(seconds) of all methods on COIL-100 object dataset.}
    \begin{tabular}{c|c|c|c|c|c|c}
    \hline
          Methods& SRRS & LRCS & RMSL & LRDE & CLRS & MR-SLMR \\
    \hline
    case 1 & 63    & \textbf{47}    & 53    & 73    & 98    & \textbf{47} \\
    case 2 & 165   & 94    & 94    & 153   & 241   & \textbf{91} \\
    case 3 & 454   & 167   & \textbf{159}   & 278   & 482   & 163 \\
    case 4 & 987   & 290   & \textbf{250}   & 517   & 823   & 270 \\
    \hline
    \end{tabular}
    \vspace{-0.2cm}
  \label{tab:time_coil}
\end{table}

\begin{figure}[htb]
    \centering
    \includegraphics[height=3.5cm,width=7cm]{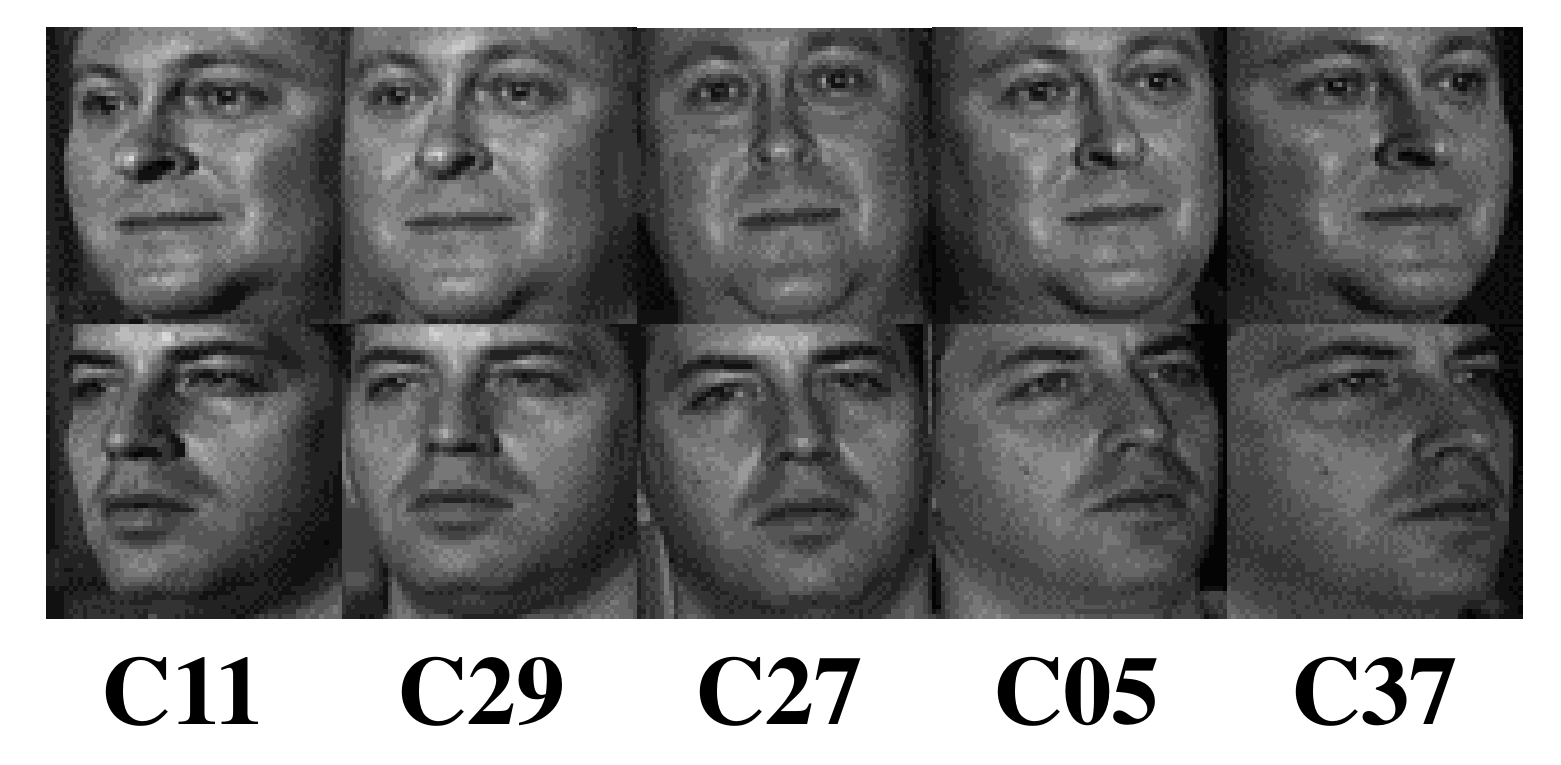}
    \caption{Exemplar subjects from CMU PIE.}
    \label{fig:img_cmu}
    \vspace{-0.2cm}
\end{figure}

\begin{figure*}[htb]
	\centering
	\begin{minipage}{3.5cm}
		\includegraphics[width=3.5cm]{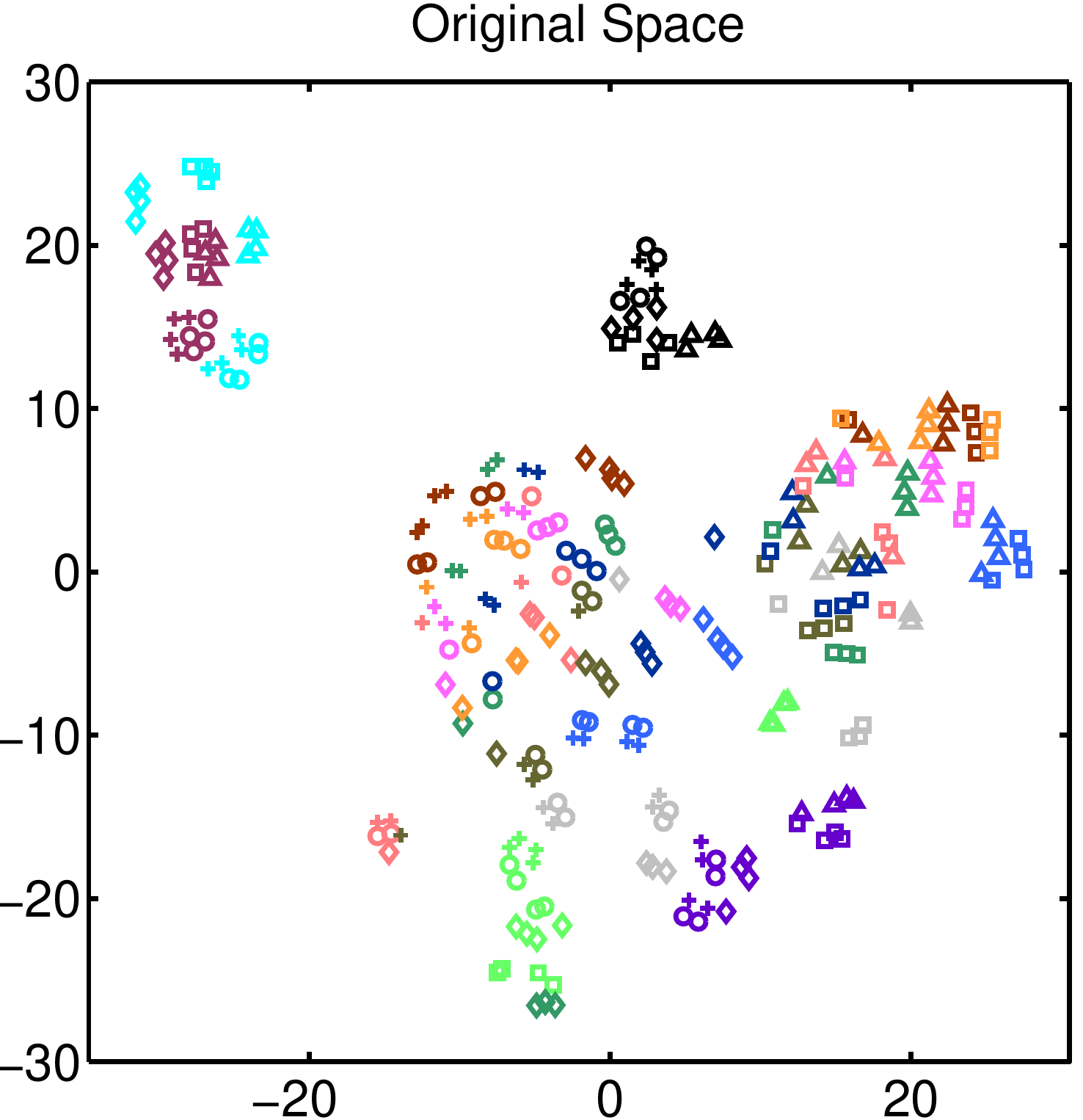}
	\end{minipage}
	\begin{minipage}{3.5cm}
		\includegraphics[width=3.5cm]{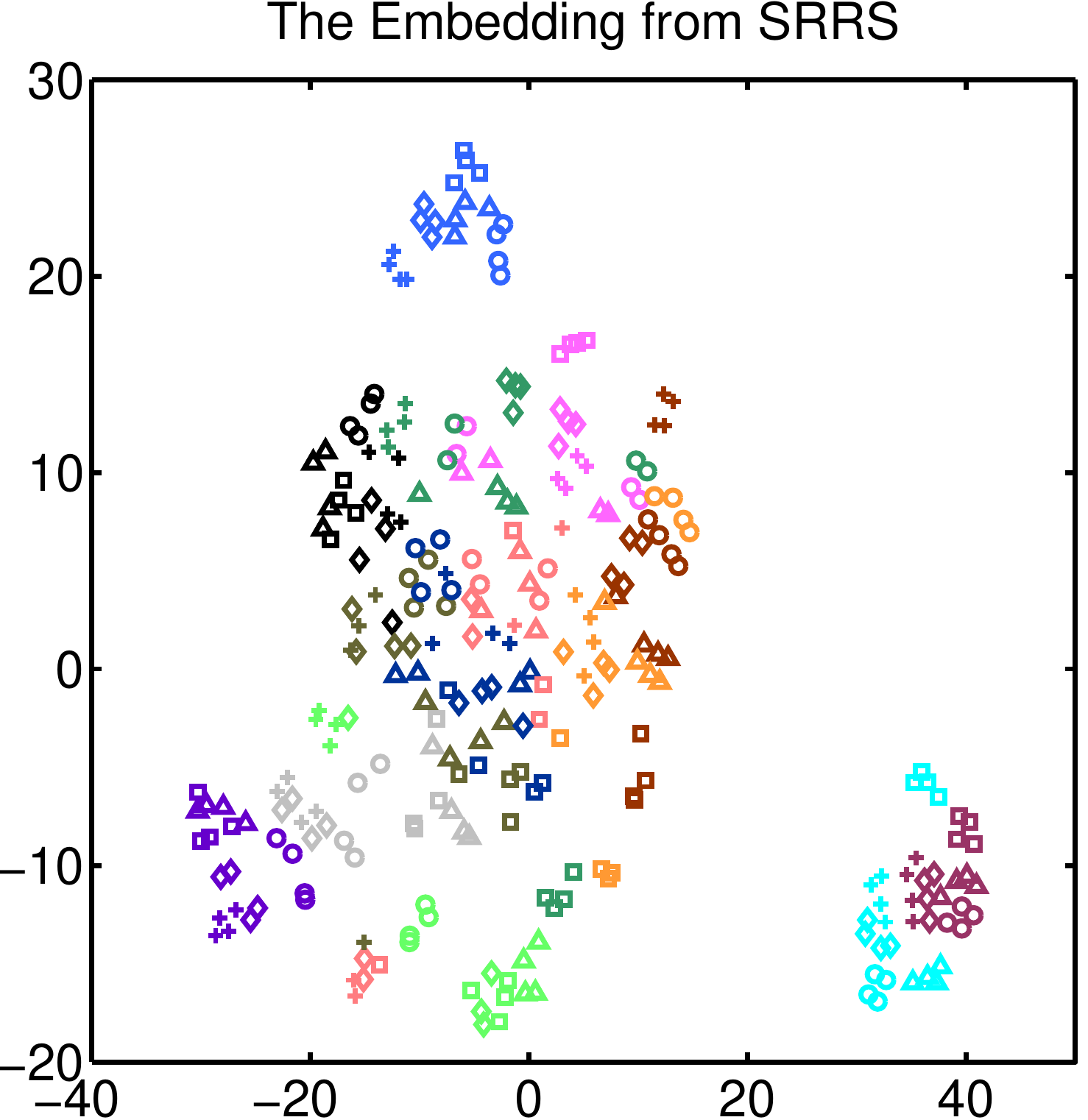}
	\end{minipage}
	\begin{minipage}{3.5cm}
		\includegraphics[width=3.5cm]{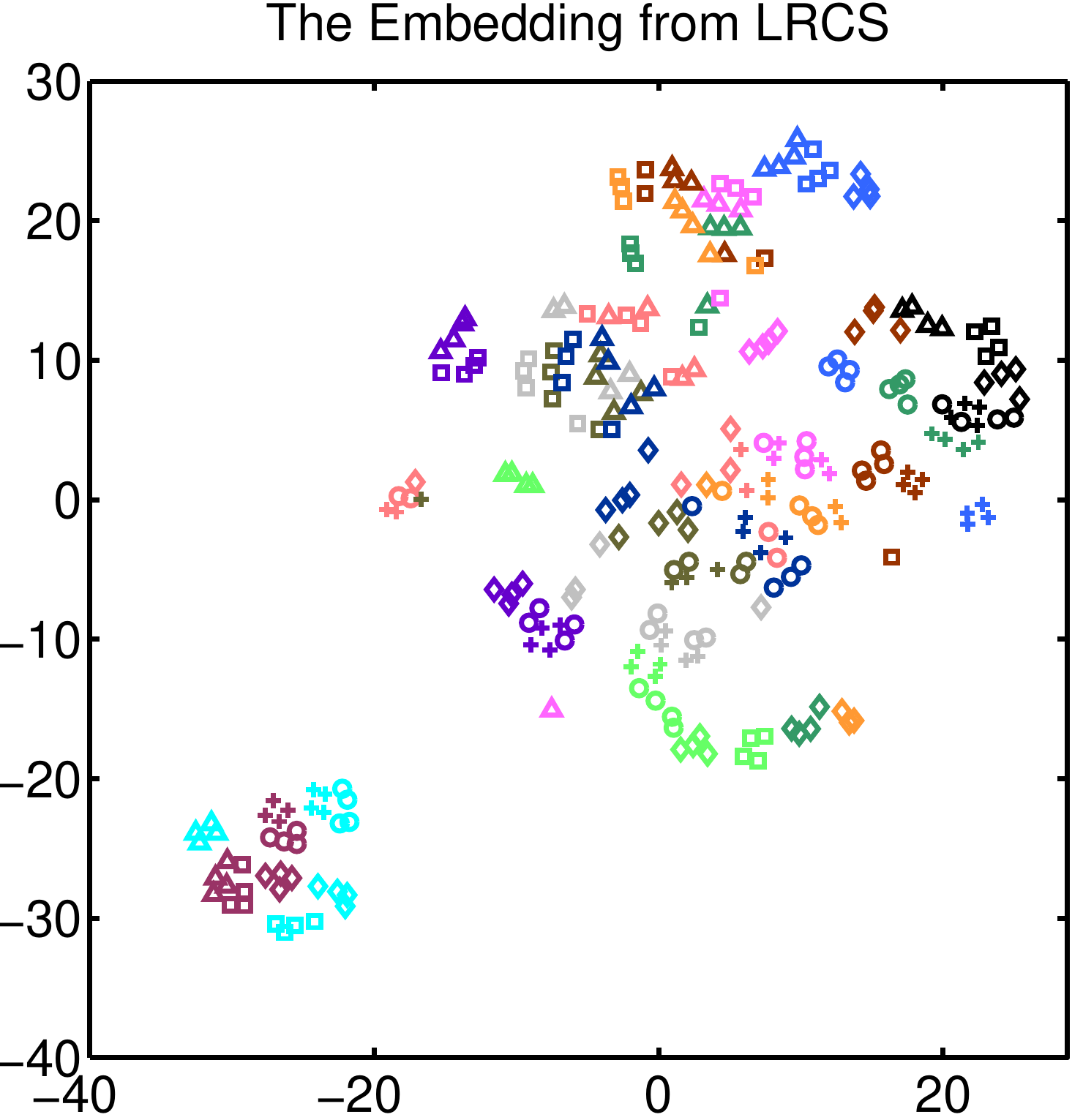}
	\end{minipage}\\
	\begin{minipage}{3.5cm}
		\includegraphics[width=3.5cm]{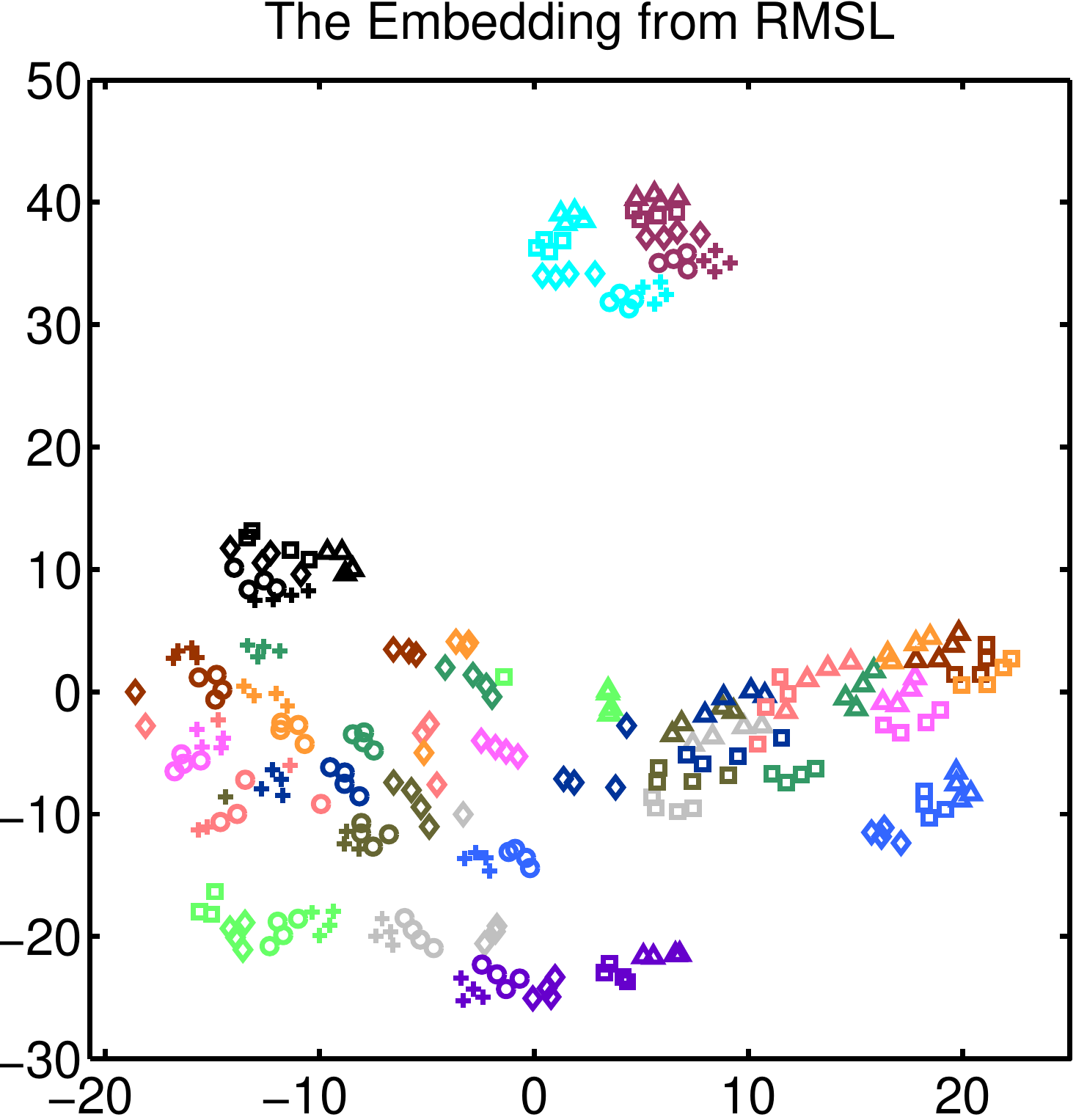}
	\end{minipage}
	\begin{minipage}{3.5cm}
		\includegraphics[width=3.5cm]{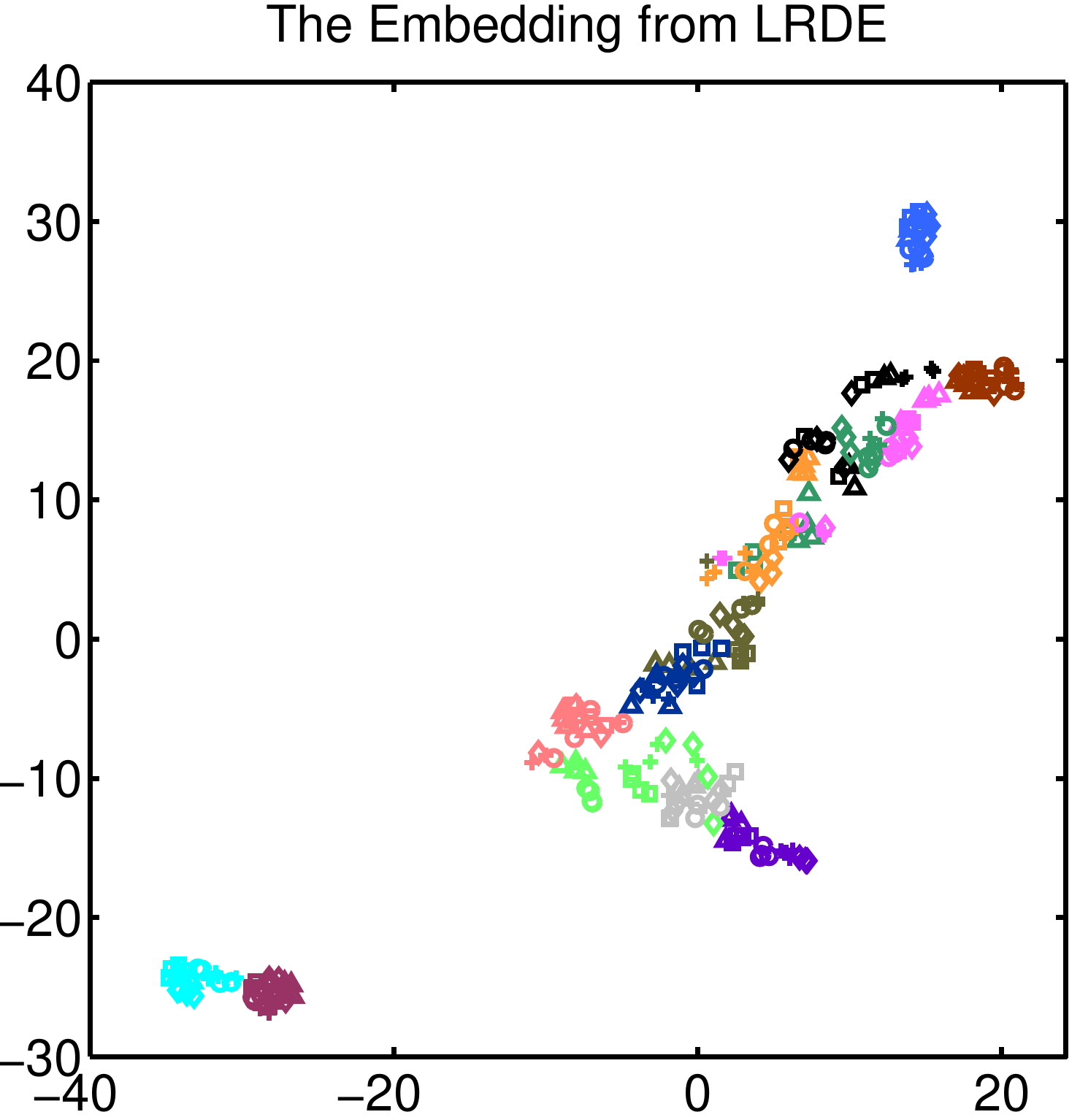}
	\end{minipage}
		\begin{minipage}{3.5cm}
		\includegraphics[width=3.5cm]{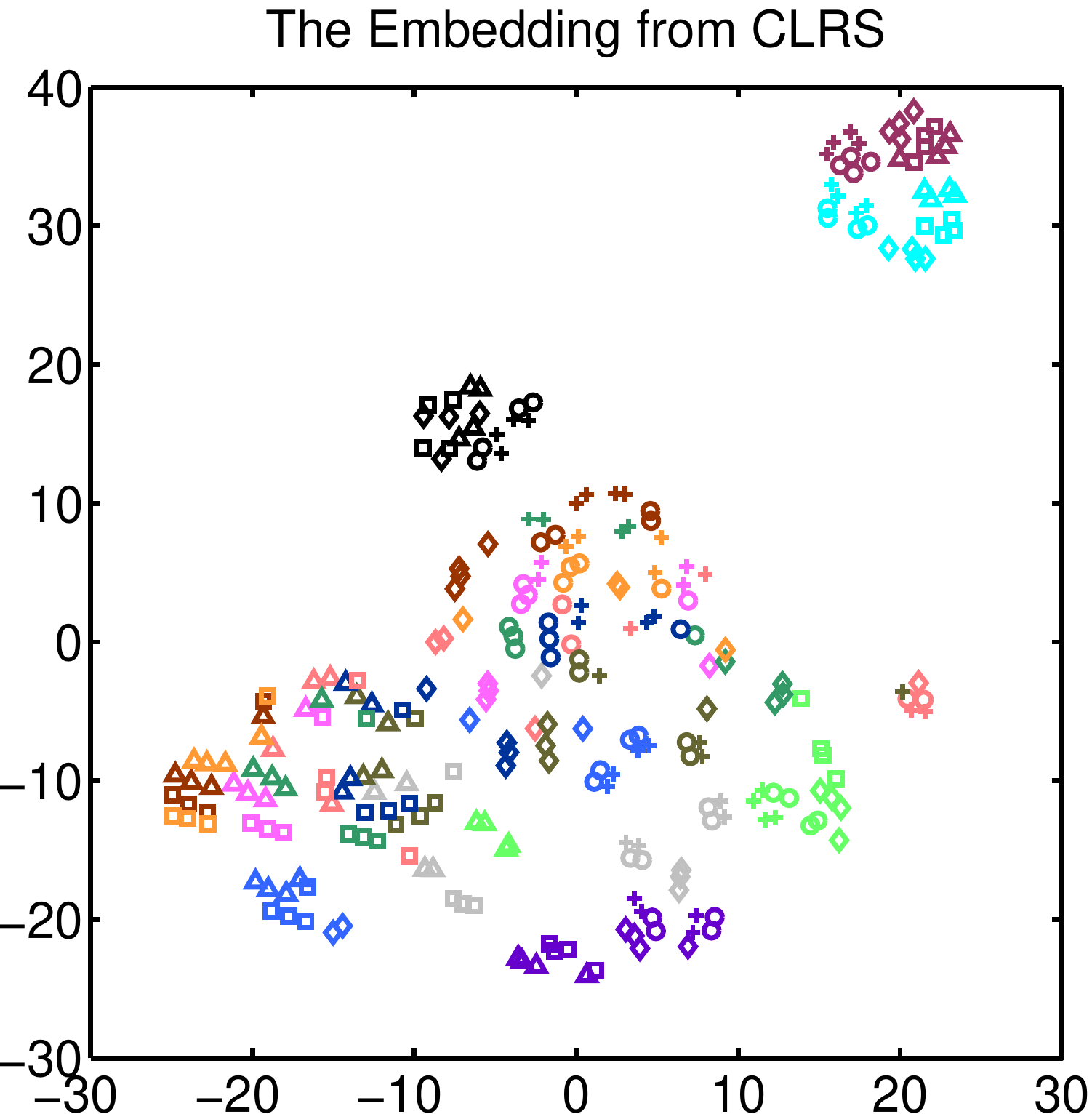}
	\end{minipage}
		\begin{minipage}{3.5cm}
		\includegraphics[width=3.5cm]{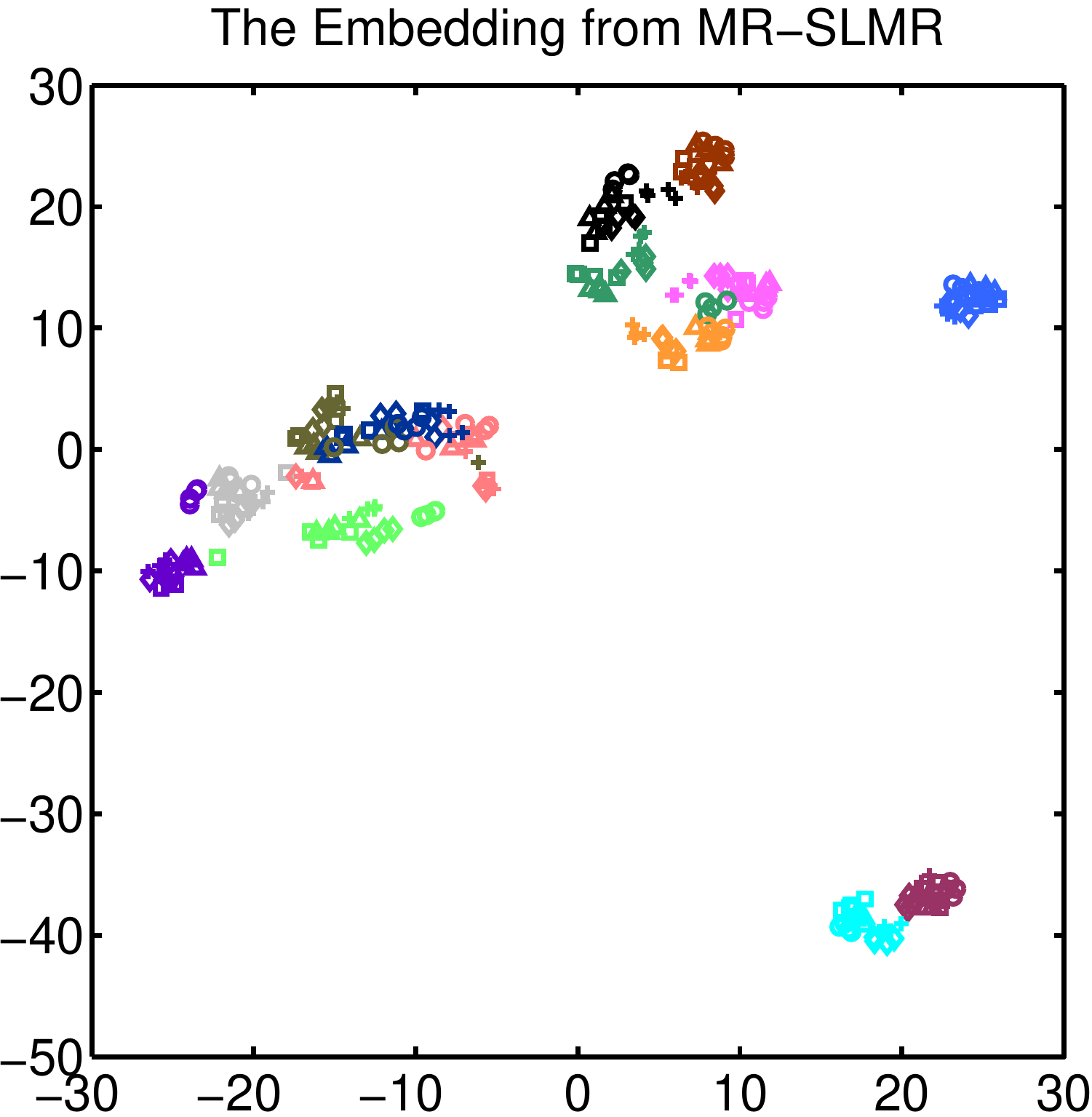}
	\end{minipage}
	\caption{Visualization of original space and the common subspace generated by LMvSL based methods on CMU PIE dataset. Markers and colors denote views and classes, respectively.}
	\vspace{-0.5cm}
	\label{cmu-vis}
\end{figure*}

\subsection{Face Recognition across Pose}
\label{subsec:face_recognition}
In this section, we demonstrate the superiority of MR-SLMR on face recognition across pose on CMU PIE~\cite{sim2002cmu}. The commonly used benchmark database contains images of 68 persons under 13 poses. As shown in Fig.~\ref{fig:img_cmu}, five poses are selected to construct multi-view data, where each person at a given pose has four cropped grayscale images of size $64\!\times\!64$. Experiments on CMU PIE are conducted in five cases to sufficiently validate the superiority of MR-SLMR in different scenario including case 1: $\{\textrm{C}27, \textrm{C}29\}$, case 2: $\{\textrm{C}27, \textrm{C}11\}$, case 3: $\{\textrm{C}05, \textrm{C}27, \textrm{C}29\}$, case 4: $\{\textrm{C}37, \textrm{C}27, \textrm{C}11\}$ and case 5: $\{\textrm{C}37, \textrm{C}05, \textrm{C}27, \textrm{C}29, \textrm{C}11\}$. In our experiments, this dataset is further divided into training set with 40 persons, validation set with 14 persons and test set with 14 persons. Furthermore, LMvSL based methods~(SRRS, LRCS, RMSL, LRDE, CLRS and SLMR-L1) and their modal regression based approaches~(term it MR-SRRS, MR-LRCS, MR-RMSL, MR-LRDE, MR-CLRS and MR-SLMR) are introduced for comparison to validate the robustness of modal regression to noise. Note that SLMR-L1 is one of our models and it can be obtained when $L_1$ norm is imposed on $\bm{E}$ in Eq.~(\ref{obj:struct2}).

\subsubsection{Superiority of MR-SLMR}
In the first experiment, the performance of LMvSL based methods is evaluated in CMU PIE. The experimental results are summarized in Table~\ref{tab:CMU_PIE_experiments_case} and the visualization of case 5 is shown in Fig.~\ref{cmu-vis}. As shown, LRCS achieves the worst performance in almost all cases of CMU PIE. Despite the consideration of discriminant information, SRRS, RMSL and CLRS perform better than LRCS with a limited improvement due to the limitation in discrepancy elimination. LRDE achieves the highest accuracy among competitors by reasons of the decrement of view discrepancy caused by criteria A, and our SLMR-L1 outperforms LRDE with a significant improvement owing to the more effective elimination of view discrepancy and enhancement of discriminability. Furthermore, although experiments are conducted on clean data, it is interesting to find that modal regression based approaches consistently outperform their baselines in almost all cases. One possible reason is that images in our experiment are cropped by ourselves, and a few images may violate cropped protocol~\cite{he2005face}. However, image pixel with a little cropped error still obeys zero-mode noise distributions. Experimental results demonstrate the superiority of MR-SLMR and the positive efficacy of modal regression. 

\begin{table}[htb]
  \centering
  \caption{Comparison results~(\%) of all methods on original CMU PIE face dataset.}
  \label{tab:CMU_PIE_experiments_case}
    \begin{tabular}{c|c|c|c|c|c}
    \hline
    Methods&case 1&case 2&case 3&csae 4&case 5\\
    \hline
    SRRS  & 74.3$\pm$4.6 & 66.8$\pm$3.1 & 69.0$\pm$3.7 & 56.2$\pm$2.4 & 59.4$\pm$2.4 \\
    MR-SRRS & 76.9$\pm$4.5 & 69.1$\pm$3.1 & 69.7$\pm$3.8 & 57.1$\pm$2.7 & 60.2$\pm$2.4 \\
    LRCS  & 74.0$\pm$4.8 & 66.1$\pm$3.1 & 68.9$\pm$3.9 & 56.3$\pm$2.2 & 58.1$\pm$2.7 \\
    MR-LRCS & 76.5$\pm$4.9 & 70.1$\pm$4.1 & 69.7$\pm$3.6 & 57.8$\pm$2.9 & 61.5$\pm$2.2 \\
    RMSL  & 75.7$\pm$5.5 & 68.0$\pm$3.4 & 70.7$\pm$3.1 & 57.9$\pm$2.3 & 62.0$\pm$2.1 \\
    MR-RMSL & 76.7$\pm$4.7 & 68.3$\pm$3.5 & 71.1$\pm$2.4 & 58.3$\pm$2.4 & 61.8$\pm$2.3 \\
    LRDE  & 81.1$\pm$4.2 & 74.1$\pm$4.6 & 84.0$\pm$3.5 & 70.3$\pm$4.0 & 79.5$\pm$3.6 \\
    MR-LRDE & 81.3$\pm$4.2 & 74.1$\pm$4.3 & 84.6$\pm$3.5 & 70.3$\pm$4.1 & 80.1$\pm$3.4 \\
    CLRS  & 75.1$\pm$4.4 & 67.3$\pm$3.1 & 69.1$\pm$3.9 & 56.7$\pm$2.7 & 60.0$\pm$2.4 \\
    MR-CLRS & 75.7$\pm$2.6 & 69.0$\pm$3.4 & 69.1$\pm$3.8 & 56.3$\pm$2.4 & 60.2$\pm$2.4 \\
    SLMR-L1 & 87.4$\pm$3.1 & 80.9$\pm$3.5 & 87.2$\pm$2.8 & 78.8$\pm$4.6 & 79.9$\pm$3.4 \\
    MR-SLMR & \textbf{88.2}$\pm$3.1 & \textbf{83.9}$\pm$4.0 & \textbf{87.7}$\pm$3.4 & \textbf{78.9}$\pm$4.1 & \textbf{80.2}$\pm$3.2 \\
    \hline
    \end{tabular}%
    \vspace{0.1cm}
\end{table}%

\begin{table*}[htb]
  \centering
  \caption{Comparison results~(\%) of all methods on corrupted CMU PIE with Gaussian noise.}
  \scalebox{0.85}{
    \begin{tabular}{c|cc|cc|cc|cc|cc|cc}
    \hline
    Power& SRRS  & MR-SRRS & LRCS  & MR-LRCS & RMSL  & MR-RMSL & LRDE  & MR-LRDE & CLRS  & MR-CLRS & SLMR-L1 & MR-SLMR \\
    \hline
    0   & 59.4(0.0) & 60.2(0.0) & 58.1(0.0) & 61.5(0.0) & 62.0(0.0) & 61.8(0.0) & 79.5(0.0) & 80.1(0.0) & 60.0(0.0) & 60.2(0.0) & 79.9(0.0) & 80.2(0.0) \\
    30  & 58.0(2.4) & 59.4(1.3) & 57.1(1.7) & 60.0(2.4) & 57.0(8.1) & 60.5(2.1) & 76.7(3.5) & 77.6(3.1) & 58.0(3.3) & 59.0(2.0) & 77.1(3.5) & 79.7(0.6) \\
    35  & 56.1(5.6) & 59.4(1.3) & 55.4(4.6) & 58.2(5.4) & 47.9(22.7) & 52.7(14.7) & 63.5(20.1) & 70.7(11.7) & 56.0(6.6) & 57.3(4.8) & 71.7(10.2) & 77.1(3.9) \\
    40  & 53.0(10.8) & 58.3(3.2) & 51.6(11.2) & 57.3(6.8) & 48.0(22.6) & 53.5(13.4) & 51.0(35.8) & 60.5(24.4) & 52.9(11.8) & 55.1(8.5) & 61.5(23.0) & 68.8(14.2) \\
    \hline
    \end{tabular}}
  \label{tab:gaussian}
\end{table*}

\begin{table*}[htb]
  \centering
  \caption{Comparison results~(\%) of all methods on corrupted CMU PIE with random noise. ``PR" denotes permutation ratio.}
  \scalebox{0.85}{
    \begin{tabular}{c|cc|cc|cc|cc|cc|cc}
    \hline
    PR&SRRS&MR-SRRS&LRCS&MR-LRCS&RMSL&MR-RMSL&LRDE&MR-LRDE&CLRS&MR-CLRS&SLMR-L1&MR-SLMR\\
    \hline
    0\%   & 59.4(0.0) & 60.2(0.0) & 58.1(0.0) & 61.5(0.0) & 62.0(0.0) & 61.8(0.0) & 79.5(0.0) & 80.1(0.0) & 60.0(0.0) & 60.2(0.0) & 79.9(0.0) & 80.2(0.0) \\
    10\%  & 58.0(2.4) & 60.2(0.0) & 58.2(-0.2) & 61.5(0.0) & 58.0(6.5) & 60.3(2.4) & 75.1(5.5) & 76.3(4.7) & 58.0(3.3) & 59.1(1.8) & 76.3(4.5) & 77.8(3.0) \\
    20\%  & 56.9(4.2) & 59.8(0.7) & 56.2(3.3) & 60.2(2.1) & 51.0(17.7) & 55.5(10.2) & 67.4(15.2) & 69.8(12.8) & 56.8(5.3) & 58.4(3.0) & 73.2(8.4) & 73.8(8.0) \\
    30\%  & 55.5(6.6) & 59.7(0.8) & 54.8(5.7) & 59.5(3.3) & 50.4(18.7) & 55.5(10.2) & 58.5(26.4) & 61.6(23.1) & 55.5(7.5) & 56.9(5.5) & 68.8(13.9) & 68.9(14.1) \\
    40\%  & 54.3(8.6) & 59.7(0.8) & 53.0(8.8) & 60.1(2.3) & 50.2(19.0) & 55.0(11.0) & 53.0(33.3) & 57.6(28.1) & 54.3(9.5) & 56.0(7.0) & 63.6(20.4) & 64.3(19.8) \\
    \hline
    \end{tabular}}
  \label{tab:random_noise}
\end{table*}

\begin{table*}[htbp]
  \centering
  \caption{Comparison results~(\%) of all methods on corrupted CMU PIE with occlusion.}
  \scalebox{0.85}{
    \begin{tabular}{c|cc|cc|cc|cc|cc|cc}
    \hline
    Area&SRRS  & MR-SRRS & LRCS  & MR-LRCS & RMSL  & MR-RMSL & LRDE  & MR-LRDE & CLRS  & MR-CLRS & SLMR-L1 & MR-SLMR \\
    \hline
    0&59.4(0.0) & 60.2(0.0) & 58.1(0.0) & 61.5(0.0) & 62.0(0.0) & 61.8(0.0) & 79.5(0.0) & 80.1(0.0) & 60.0(0.0) & 60.2(0.0) & 79.9(0.0) & 80.2(0.0) \\
    16&60.0(-1.0) & 61.2(-1.7) & 58.6(-0.9) & 61.6(-0.2) & 61.8(0.3) & 61.7(0.0) & 70.5(11.3) & 72.3(9.7) & 59.8(0.3) & 60.6(-0.7) & 74.5(6.8) & 75.5(5.9) \\
    24&60.3(-1.5) & 60.6(-0.7) & 58.3(-0.3) & 61.7(-0.3) & 60.9(1.8) & 61.3(0.7) & 66.5(16.4) & 67.9(15.2) & 60.2(-0.3) & 60.6(-0.7) & 70.8(11.4) & 72.7(9.4) \\
    32&59.5(-0.2) & 60.4(-0.3) & 58.7(-1.0) & 61.2(0.5) & 56.8(8.4) & 61.1(1.0) & 65.4(17.7) & 67.0(16.4) & 59.2(1.3) & 60.2(0.0) & 68.4(14.4) & 71.4(11.0) \\
    48&57.8(2.7) & 59.0(2.0) & 56.3(3.1) & 60.6(1.5) & 56.5(8.9) & 59.2(4.1) & 66.8(16.0) & 68.2(14.9) & 57.7(3.8) & 58.9(2.2) & 67.3(15.8) & 68.4(14.7)\\
    \hline
    \end{tabular}}
  \label{tab:occl}
\end{table*}

% Table generated by Excel2LaTeX from sheet 'CMU PIE'
\begin{table*}[htb]
  \centering
  \caption{Comparison results~(\%) of all methods on corrupted CMU PIE with outliers. ``PR" denotes permutation ratio.}
  \scalebox{0.85}{
    \begin{tabular}{c|cc|cc|cc|cc|cc|cc}
    \hline
    PR& SRRS & MR-SRRS & LRCS & MR-LRCS & RMSL & MR-RMSL & LRDE & MR-LRDE & CLRS & MR-CLRS & SLMR-L1 & MR-SLMR \\
    \hline
    0\%   & 59.4(0.0) & 60.2(0.0) & 58.1(0.0) & 61.5(0.0) & 62.0(0.0) & 61.8(0.0) & 79.5(0.0) & 80.1(0.0) & 60.0(0.0) & 60.2(0.0) & 79.9(0.0) & 80.2(0.0) \\
    5\%   & 59.1(0.5) & 60.0(0.3) & 56.4(2.9) & 58.7(4.5) & 59.6(3.9) & 59.8(3.2) & 75.5(5.0) & 76.0(5.1) & 60.0(0.0) & 60.1(0.2) & 74.9(6.3) & 77.7(3.1) \\
    10\%  & 59.0(0.7) & 59.6(1.0) & 55.2(5.0) & 59.1(3.9) & 55.7(10.2) & 59.6(3.6) & 70.3(11.6) & 71.0(11.4) & 59.8(0.3) & 59.9(0.5) & 72.7(9.0) & 76.4(4.7) \\
    15\%  & 58.4(1.7) & 59.2(1.7) & 55.2(5.0) & 59.0(4.1) & 46.4(25.2) & 59.6(3.6) & 68.5(13.8) & 69.0(13.9) & 59.8(0.3) & 59.8(0.7) & 68.9(13.8) & 74.3(7.4) \\
    20\%  & 58.5(1.5) & 59.0(2.0) & 54.7(5.9) & 59.2(3.7) & 44.2(28.7) & 59.4(3.9) & 66.2(16.7) & 66.4(17.1) & 59.6(0.6) & 59.7(0.8) & 60.8(23.9) & 72.1(10.1) \\
    \hline
    \end{tabular}}%
  \label{tab:outlier_noise}%
\end{table*}%
	
\begin{figure*}[!htb]
	\centering
	\begin{minipage}[t]{4cm}
		\includegraphics[width=4cm]{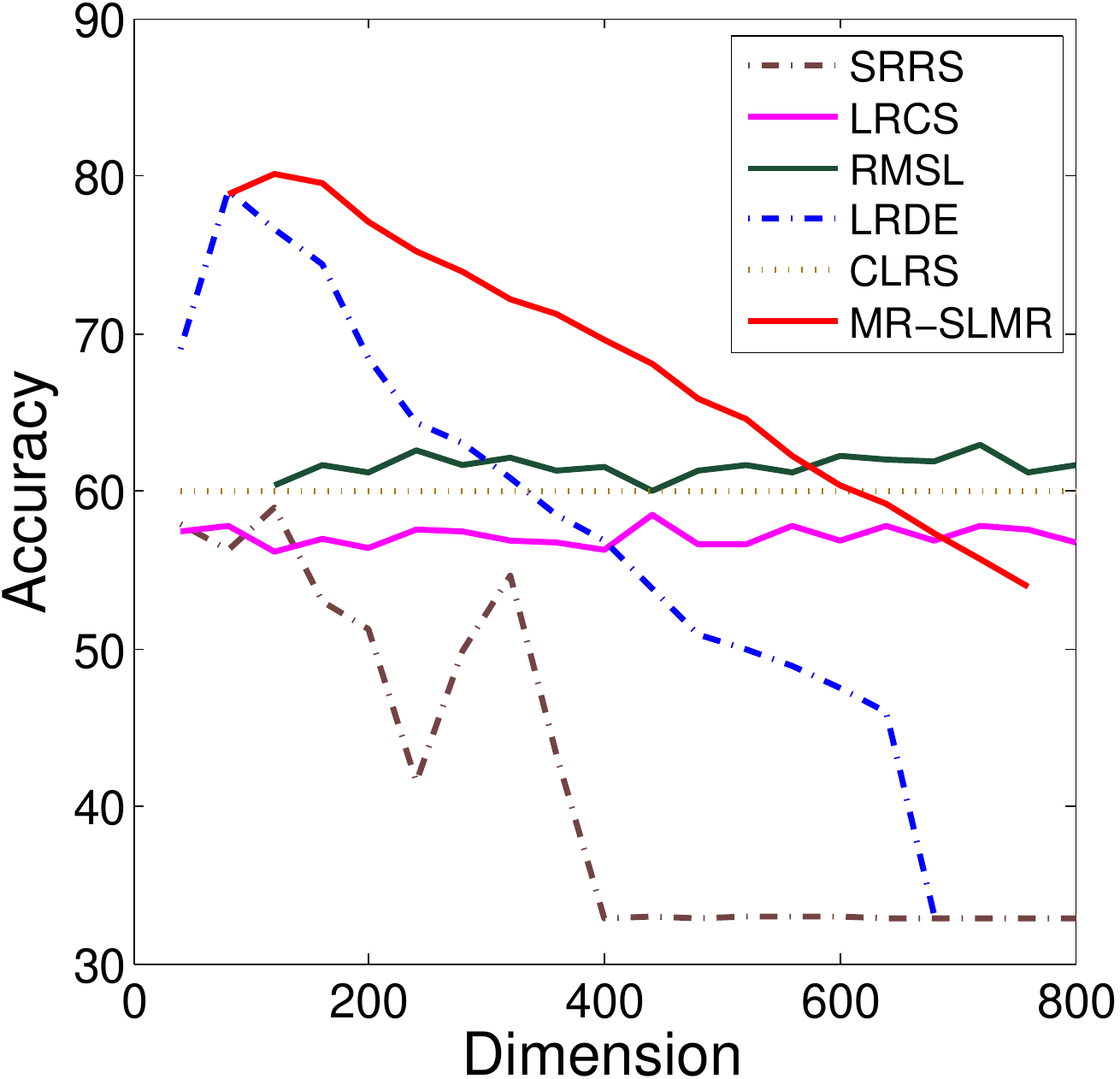}
		\subcaption{}
	\end{minipage}
	\begin{minipage}[t]{4cm}
		\includegraphics[width=3.95cm]{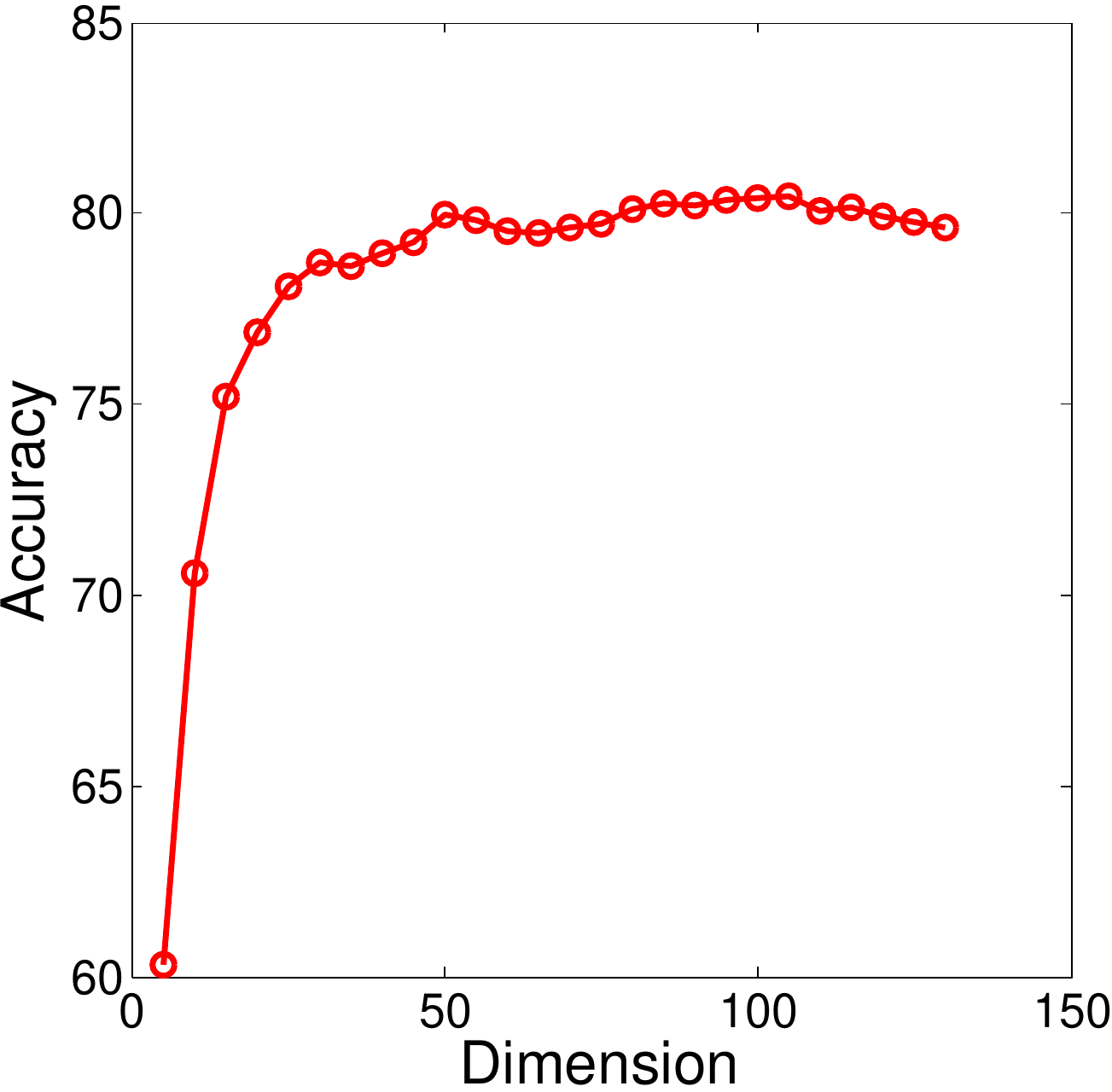}
		\subcaption{}
	\end{minipage}
	\begin{minipage}[t]{4cm}
		\includegraphics[width=4cm]{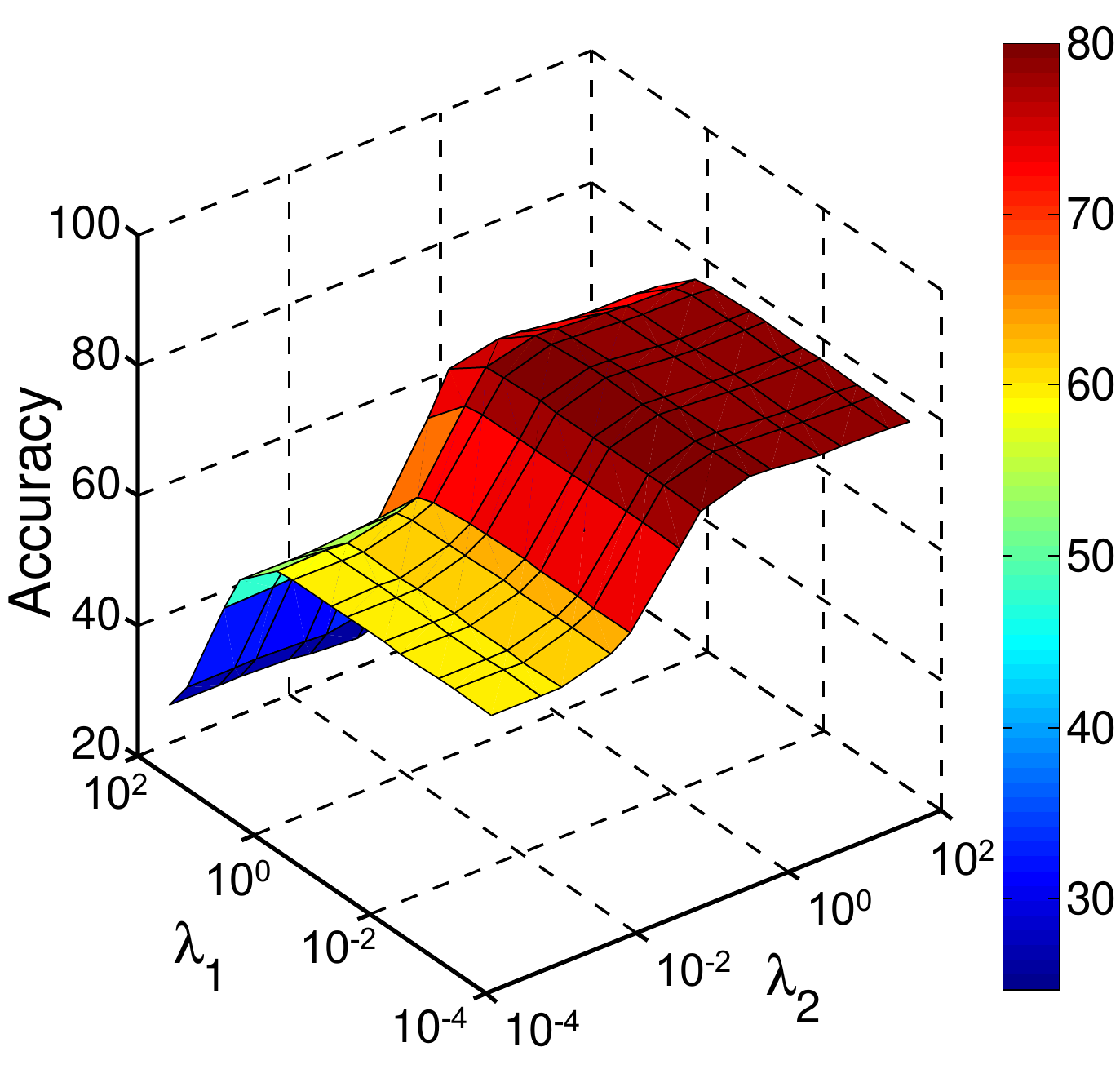}
		\subcaption{}
	\end{minipage}%
	\hspace{0.4cm}
	\begin{minipage}[t]{4cm}
		\includegraphics[width=4cm]{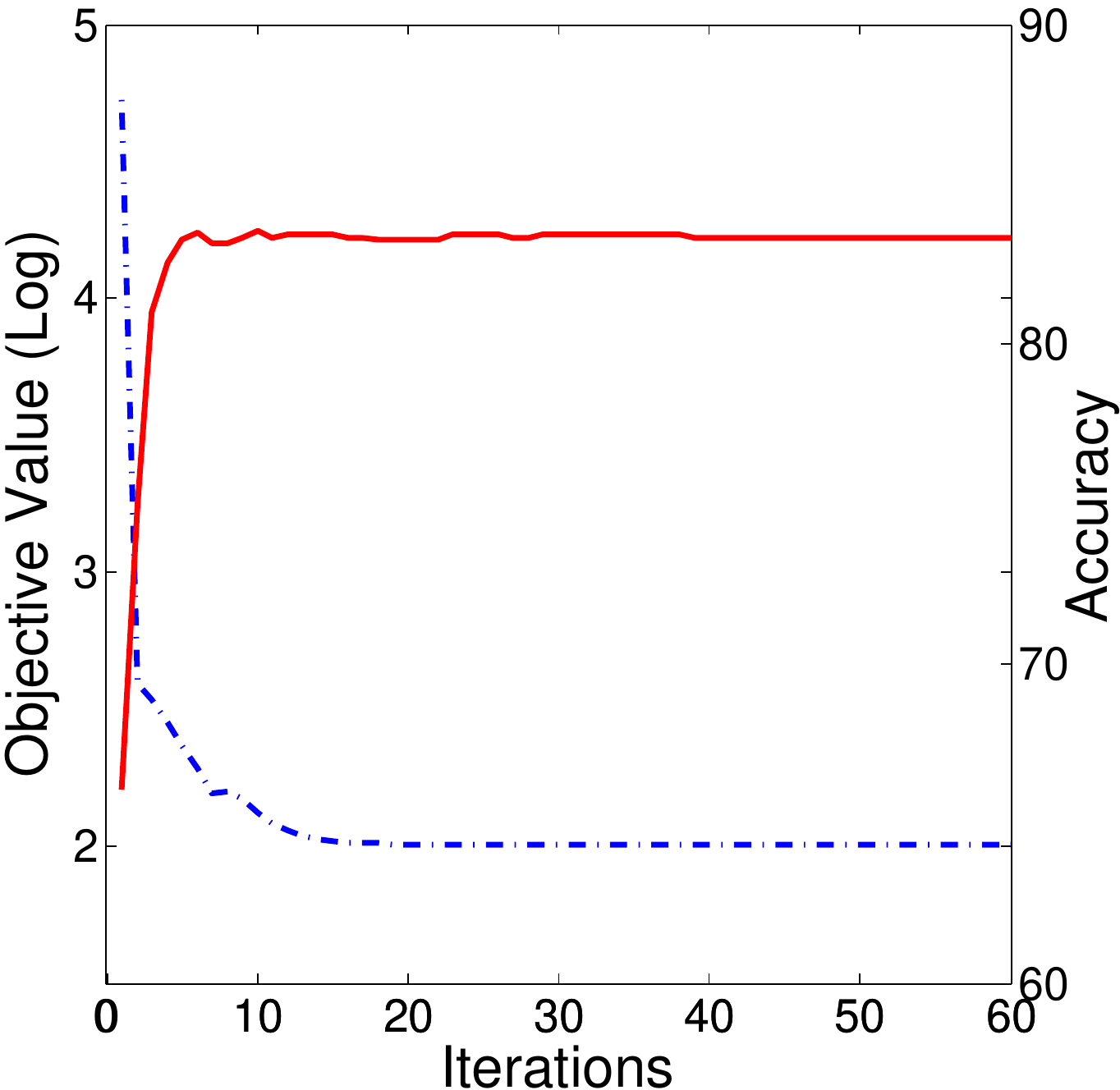}
		\subcaption{}
	\end{minipage}
	\caption{Performance of LMvSL based methods in case 5 of original CMU PIE. (a) Accuracy of all competitors with different PCA dimensions. (b) Accuracy of our MR-SLMR with different subspace dimensions. (c) Accuracy of our MR-SLMR with different hyper-parameter values\{$\lambda_1$, $\lambda_2$\}. (d) Convergence curve~(\textcolor{blue}{blue}) and accuracy curve~(\textcolor{red}{red}) of our MR-SLMR.}
	\label{fig:curve}
	\vspace{-0.5cm}
\end{figure*}

\subsubsection{Robustness Analysis of MR-SLMR}
In the second experiment, we evaluate the robustness of modal regression based methods to different zero-mode noise including Gaussian noise, random noise, occlusion and outliers, where Gaussian white noise of $30$dBW, $35$dBW and $40$dBW power are respectively added to clean images, random noise is added to clean images by replacing $10\%$, $20\%$, $30\%$ and $40\%$ pixels with values at $\left[0, 255\right]$, occlusion noise is added by overlapping clean image with a given picture of size $16\!\times\!16$, $24\!\times\!24$, $32\!\times\!32$ and $48\!\times\!48$ and outliers are added by replacing $5\%$, $10\%$, $15\%$ and $20\%$ images with random values. Experimental results are presented in Table~\ref{tab:gaussian}, \ref{tab:random_noise}, \ref{tab:occl} and \ref{tab:outlier_noise} respectively, where values in parentheses stand for the relative performance loss $\left(\%\right)$ with respect to the scenario in the clean images. As shown in Table \ref{tab:gaussian}, \ref{tab:random_noise} and \ref{tab:occl}, modal regression based methods are much more insensitive to these noise than their corresponding baselines. Taking permutation scenario with $40\%$ random noise as an example, the accuracy of MR-SRRS decreases to $59.7\%$ with a relative $0.8\%$ performance drop, whereas SRRS suffers from a performance drop near $8.6\%$ from its original $59.4\%$ accuracy. It is worth noting that similar accuracy is obtained for SLMR-L1 and MR-SLMR due to the fact that sparse regularization is used in SLMR-L1 algorithm. Furthermore, as can be seen in Table \ref{tab:outlier_noise}, it is interesting to find that SRRS, LRDE and CLRS are even slightly more robust to outliers. One possible reason is that $\ell_{2,1}$ regularization used in SRRS, LRDE and CLRS can handle outliers more effectively.%effectively　

\subsection{Property Analysis}
\label{subsec:property_analysis}
In this section, experiments to evaluate the influence of PCA dimensions, subspace dimensions and hyper-parameters on the classification performance of LMvSL based algorithms are conducted in case 5 of CMU PIE, where PCA dimensions are traversed from 40 to 800 at an interval of 40, subspace dimensions are traversed from 5 to 130 at an interval of 5 and hyper-parameters, namely $\lambda_1$ and $\lambda_2$, are both in range $\left[1\rm{e}\!-\!4, 5\rm{e}\!-\!4, 1\rm{e}\!-\!3, 5\rm{e}\!-\!3, 1\rm{e}\!-\!2, 5\rm{e}\!-\!2, 0.1, 0.5, 1, 5, 10, 100\right]$. Experimental results are shown in Fig.~\ref{fig:curve} (a), (b) and (c). 

As can be seen, the performance of LRCS, RMSL and CLRS do not suffer from the change of PCA dimensions compared with SRRS and MR-SLMR, whereas LRDE and our MR-SLMR achieve satisfactory accuracy in a low PCA dimension. By contrast, our MR-SLMR is insensitive to subspace dimensions. This nice property can benefit real applications. Furthermore, the accuracy of MR-SLMR varies dramatically with $\lambda_1$ and $\lambda_2$ in a relatively large range. We suggest using cross validation to determine their values in real applications.

Finally, we conduct convergence analysis in case 5 of CMU PIE. The changing trends of loss and accuracy with respect to iterations are presented in Fig.~\ref{fig:curve} (d). As can be seen, our MR-SLMR achieves the highest performance and its loss reaches minimum after a few iterations. This indicates that the optimization method can lead to a fast convergence.

\section{Conclusion}
\label{sec:conclusion}
In this paper, inspired by the block-diagonal representation learning and modal regression, we present a novel Modal Regression based Structured Low-rank Matrix Recovery for cross-view classification. Extensive experiments on four commonly used datasets demonstrate that our method can more effectively remove view discrepancy and improve discriminancy simultaneously compared with other state-of-the-art methods, which thus results in a satisfactory performance under the same scale of computational complexity, and robustness analysis on CMU PIE indicates the positive efficacy of modal regression on handling complicated noise. 

From the in-depth analysis in Sect.~\ref{subsec:handwritten_digits}, our method based on block-diagonal representation learning presents inferiority in improving intra-view discriminancy compared with the framework of graph embedding. A feasible method to handle this problem is to integrate more proper discriminant regularization. Hence, we are interested in developing a more discriminative model in future work.

\section*{Acknowledgment}
This work was supported partially by the National Natural Science Foundation of China~(No. 11671161, 61571205 and 61772220), the Key Program for International S\&T Cooperation Projects of China~(No. 2016YFE0121200), the Special Projects for Technology Innovation of Hubei Province~(No. 2018ACA135), the Key Science and Technology Innovation Program of Hubei Province~(No. 2017AAA017), the Natural Science Foundation of Hubei Province~(No. 2018CFB691), fund from Science, Technology and Innovation Commission of Shenzhen Municipality~(No. JCYJ20180305180637611, JCYJ20180305180804836 and JSGG20180507182030600).
%This work was supported partially by the Key Science and Technology of Shenzhen~(No. CXZZ20150814155434903), in part by the Key Program for International S\&T Cooperation Projects of China~(No. 2016YFE0121200), in part by the Key Science and Technology Innovation Program of Hubei Province~(No. 2017AAA017), in part by the Special Projects for Technology Innovation of Hubei Province~(No. 2018ACA135), in part by the Shenzhen Research Council(No. JCYJ20180305180804836), in part by the National Natural Science Foundation of China~(No. 61571205 and 61772220).

\ifCLASSOPTIONcaptionsoff
  \newpage
\fi

{
\bibliographystyle{IEEEtran}
\bibliography{egbib.bib}
}

\end{document}